\documentclass{preprint}
\usepackage[T1]{fontenc}
\usepackage{amsmath} 
\usepackage[table]{xcolor}
\usepackage{booktabs}
\usepackage{graphicx} 
\usepackage{mdframed}
\usepackage{xspace}
\usepackage{makecell}
\usepackage{tabularx}
\usepackage{bbm}
\usepackage{listings}
\usepackage{subcaption}
\usepackage{setspace}
\usepackage{multirow}
\usepackage{enumitem}
\usepackage{tabularray}
\usepackage[colorlinks=true]{hyperref}
\usepackage[numbers,comma,sort&compress]{natbib} 
\usepackage{xurl}

\colorlet{punct}{red!60!black}
\definecolor{background}{HTML}{EEEEEE}
\definecolor{delim}{RGB}{20,105,176}
\colorlet{numb}{magenta!60!black}
\lstdefinelanguage{json}{
    basicstyle=\footnotesize\ttfamily,
    numberstyle=\scriptsize,
    stepnumber=1,
    numbersep=8pt,
    showstringspaces=false,
    breaklines=true,
    upquote=true,
    literate=
     *{0}{{{\color{numb}0}}}{1}
      {1}{{{\color{numb}1}}}{1}
      {2}{{{\color{numb}2}}}{1}
      {3}{{{\color{numb}3}}}{1}
      {4}{{{\color{numb}4}}}{1}
      {5}{{{\color{numb}5}}}{1}
      {6}{{{\color{numb}6}}}{1}
      {7}{{{\color{numb}7}}}{1}
      {8}{{{\color{numb}8}}}{1}
      {9}{{{\color{numb}9}}}{1}
      {:}{{{\color{punct}{:}}}}{1}
      {,}{{{\color{punct}{,}}}}{1}
      {\{}{{{\color{delim}{\{}}}}{1}
      {\}}{{{\color{delim}{\}}}}}{1}
      {[}{{{\color{delim}{[}}}}{1}
      {]}{{{\color{delim}{]}}}}{1},
}
\lstdefinelanguage{prompt}{
    breaklines=true,
    breakindent=0pt,
    basicstyle=\footnotesize\ttfamily,
    columns=fullflexible,
    upquote=true,
    identifierstyle=,
}

\definecolor{graybg}{HTML}{E2E2EB}
\renewcommand{\emph}[1]{\textbf{\textit{#1}}}

\newcommand\boxedname{Prompt\xspace}

\newcounter{prompt}
\newenvironment{prompt}[1][]{
   \refstepcounter{prompt}
    \begin{mdframed}[
        innertopmargin=2pt, 
        innerbottommargin=2pt,
        innerleftmargin=2pt, 
        innerrightmargin=2pt,
        frametitle={\textbf{\boxedname~\theprompt:} #1},
        frametitlefont=\footnotesize,
        frametitlerule=true,
        backgroundcolor=graybg]%
    \setlength{\parindent}{0pt}%
    \setlength{\parskip}{.5em}%
    \scriptsize\ttfamily\hyphenchar\font=`\-\spaceskip=.5em plus .5em\xspaceskip=.5em%
}
{%
    \par%
    \end{mdframed}%
}

\title{Reinforcement Learning Improves LLM Accuracy and Reasoning in Disease Classification from Radiology Reports}
\author[1,2]{Yishu Wei}
\author[1]{Yi Lin}
\author[3]{Adam Flanders}
\author[2]{George Shih}
\author[1,2,*]{Yifan Peng}
\affil[1]{Department of Population Health Sciences, Weill Cornell Medicine, New York, NY}
\affil[2]{Department of Radiology, Weill Cornell Medicine, New York, NY}
\affil[3]{Department of Radiology, Thomas Jefferson University, Philadelphia, PA}
\affil[*]{Corresponding: \url{yip4002@med.cornell.edu}}

\begin{document}

\maketitle

\begin{abstract}
Accurate disease classification from radiology reports is essential for many applications. While supervised fine-tuning (SFT) of lightweight LLMs improves accuracy, it can degrade reasoning. We propose a two-stage approach: SFT on disease labels followed by Group Relative Policy Optimization (GRPO) to refine predictions by optimizing accuracy and format without reasoning supervision. Across three radiologist-annotated datasets, SFT outperformed baselines and GRPO further improved classification and enhanced reasoning recall and comprehensiveness.
\end{abstract}

\section*{Main Text}\label{introduction}

Recent advances in deep learning-driven computer-aided diagnosis have underscored the crucial role of high-quality and diverse medical imaging datasets in model development. However, such datasets remain scarce, particularly with respect to comprehensive annotation sets that provide expert-curated labels of abnormal findings in radiographs. The generation of these annotations has traditionally been both labor-intensive and time-consuming.

To address this bottleneck, increasing attention has been directed toward natural language processing (NLP) and, more recently, large language models (LLMs), as a means of automatically extracting and categorizing diagnostic information from free-text radiology reports \cite{jain2021radgraph, chambon2024chexpert}. 
However, the adoption of LLMs in disease classification remains constrained by practical considerations. Using external Application Programming Interfaces (APIs) reduces the need for specialized on-site hardware, but raises significant concerns about data privacy and security. In contrast, local hosting mitigates privacy risks but often demands substantial computational resources. In this context, lightweight LLMs offer a promising compromise: their smaller parameter counts facilitate on-premise deployment and enable fine-tuning on limited, task-specific datasets to achieve optimal performance in clinical settings \cite{Tran2024-wm, Garg2025-hs}.

Supervised fine-tuning (SFT) has emerged as a prominent approach for optimizing the performance of lightweight LLMs \cite{Ben-Shoham2024-ha, Jiang2024-ze}. In disease classification tasks, SFT has been shown to allow lightweight LLMs to achieve accuracies comparable to those of larger and more resource-intensive models \cite{Wei2024-dm}. However, SFT also introduces a key limitation: it may degrade the model's ability to generate explicit and coherent reasoning paths, in part because fine-tuning typically omits detailed reasoning or intermediate annotation labels. This limitation is particularly consequential in medical domains, such as radiology, where transparency and explainability are crucial for clinical decision-making and for fostering trust in algorithmic recommendations. Insufficient reasoning capabilities thus represent a significant barrier to the broader adoption of LLMs in the radiology domain.

\begin{figure}
\centering
\includegraphics[width=\linewidth]{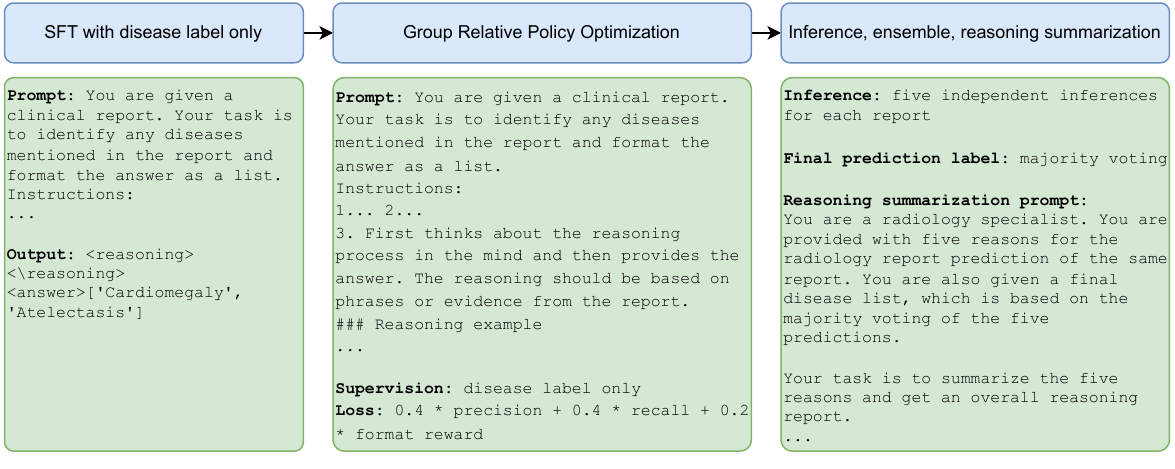}
\caption{Two-stage training and inference workflow.}
\label{fig:workflow}
\end{figure}

To overcome this challenge, we propose a multi-stage training and inference framework to improve the reasoning capabilities of lightweight LLMs for disease classification from radiology reports (Figure~\ref{fig:workflow}). In the first stage, SFT is performed using only disease labels, allowing the model to adapt to the radiology domain under outcome supervision. In the second stage, we apply a reinforcement learning algorithm (i.e., Group Relative Policy Optimization [GRPO]) on the same dataset \textit{without} requiring explicit reasoning supervision \cite{DeepSeek-AI2025-pz, Setlur2024-zp, Havrilla2024-dn}. Instead, we used a rule-based reward function to verify both the \textbf{classification accuracy} and \textbf{formatting correctness}. In particular, the formatting reward assigns a value of zero when no valid reasoning can be extracted in the specified output format, thereby encouraging the model to generate explicit reasoning alongside disease labels.

During inference, the framework generates multiple outputs for each radiology report, each comprising both diseases and corresponding reasoning. Final disease labels are determined via majority voting, thereby improving the classification robustness and reliability. To consolidate the reasoning component, the original (unfine-tuned) lightweight LLM is prompted to synthesize all generated explanations into a single, coherent rationale. 

In this study, we utilized three datasets of de-identified, structured radiology reports: MIMIC-CXR, NIH-CXR \cite{Wang2017-lm}, and MIDRC  \cite{Zhou2024-ib} (Table~\ref{tab:data}). For all evaluation datasets, radiologists were instructed to re-examine the entire report and list all disease labels that they could identify. 
In addition, we evaluated three lightweight LLMs capable of local deployment: LLaMA 3.1-8B-Instruct (LLaMa) \cite{Grattafiori2024-rd}, Qwen 2.5-3B-Instruct (Qwen) \cite{Qwen2024-mg}, and Phi-3 Min-128K-Instruct (Phi3) \cite{Abdin2024-uo}. 
Therefore, across the three gold-standard datasets, a total of nine evaluation cohorts were defined. 
\begin{table}
\caption{Statistics of datasets. card. - cardiomediastinum.}
\label{tab:data}
\centering
\small
\begin{tabular}{lcccccc}
\toprule
& \multicolumn{4}{c}{MIMIC-CXR}\\
\cmidrule{2-5}
Disease & Training & Reasoning & Evaluation & \makecell[c]{Eval-Silver} & NIH-CXR & MIRDC \\ 
\midrule
Usage & \makecell[c]{Fine-tuning \\\& GRPO training} & \makecell[c]{Fine-tuning \\(reasoning)} & Evaluation & Evaluation & Evaluation & Evaluation \\ 
Annotator & GPT4-o & GPT4-o & Radiologists & GPT4-o & Radiologists & Radiologists \\ 
Reports & 2,000 & 200 & 100 & 1,000 & 50 & 50 \\ 
Average length & 47.9 & 45.8 & 60 & 48.1 & 52 & 58.3 \\ 
Average labels & 1.59 & 1.65 & 1.93 & 1.61 & 2.52 & 2.59 \\ 
Disease & ~ & ~ & ~ & ~ & ~ & ~ \\ 
~~~~Atelectasis & 0.203 & 0.230 & 0.270 & 0.217 & 0.120 & 0.000 \\ 
~~~~Cardiomegaly & 0.176 & 0.215 & 0.350 & 0.198 & 0.010 & 0.060 \\ 
~~~~Consolidation & 0.035 & 0.015 & 0.020 & 0.033 & 0.040 & 0.020 \\ 
~~~~Edema & 0.084 & 0.090 & 0.180 & 0.088 & 0.060 & 0.000 \\ 
~~~~Enlarged card. & 0.011 & 0.010 & 0.010 & 0.010 & 0.000 & 0.000 \\ 
~~~~Fracture & 0.031 & 0.040 & 0.060 & 0.027 & 0.020 & 0.000 \\ 
~~~~Lung lesion & 0.009 & 0.005 & 0.000 & 0.011 & 0.020 & 0.000 \\ 
~~~~Lung opacity & 0.173 & 0.150 & 0.330 & 0.196 & 0.760 & 0.980 \\ 
~~~~Pleural effusion & 0.201 & 0.215 & 0.310 & 0.205 & 0.620 & 0.780 \\ 
~~~~Pleural other & 0.006 & 0.010 & 0.000 & 0.003 & 0.000 & 0.000 \\ 
~~~~Pneumonia & 0.033 & 0.020 & 0.050 & 0.028 & 0.020 & 0.020 \\ 
~~~~Pneumothorax & 0.025 & 0.040 & 0.030 & 0.025 & 0.100 & 0.020 \\ 
~~~~Support devices & 0.200 & 0.205 & 0.250 & 0.212 & 0.680 & 0.700 \\ 
\bottomrule
\end{tabular}
\end{table}

For the disease classification task, we evaluated performance using micro-averaged precision, recall, and F1 score (detailed in Section Evaluation metrics). While multiple inference is a significant contribution of our framework, we applied the same ensemble of multiple predictions to the baseline and SFT models to rule out its impact. Ensembling almost consistently improves the performance of baseline and SFT as well (Supplementary Table \ref{stab:individual_mimic}).
To assess the quality of generated reasoning, we propose two metrics: \textbf{reasoning recall}, which measures the proportion of ground-truth diseases referenced in the reasoning, and \textbf{reasoning comprehensiveness}, which quantifies the proportion of predicted diseases that are supported by explicit reasoning.

\begin{figure}
\centering
\includegraphics[width=\linewidth]{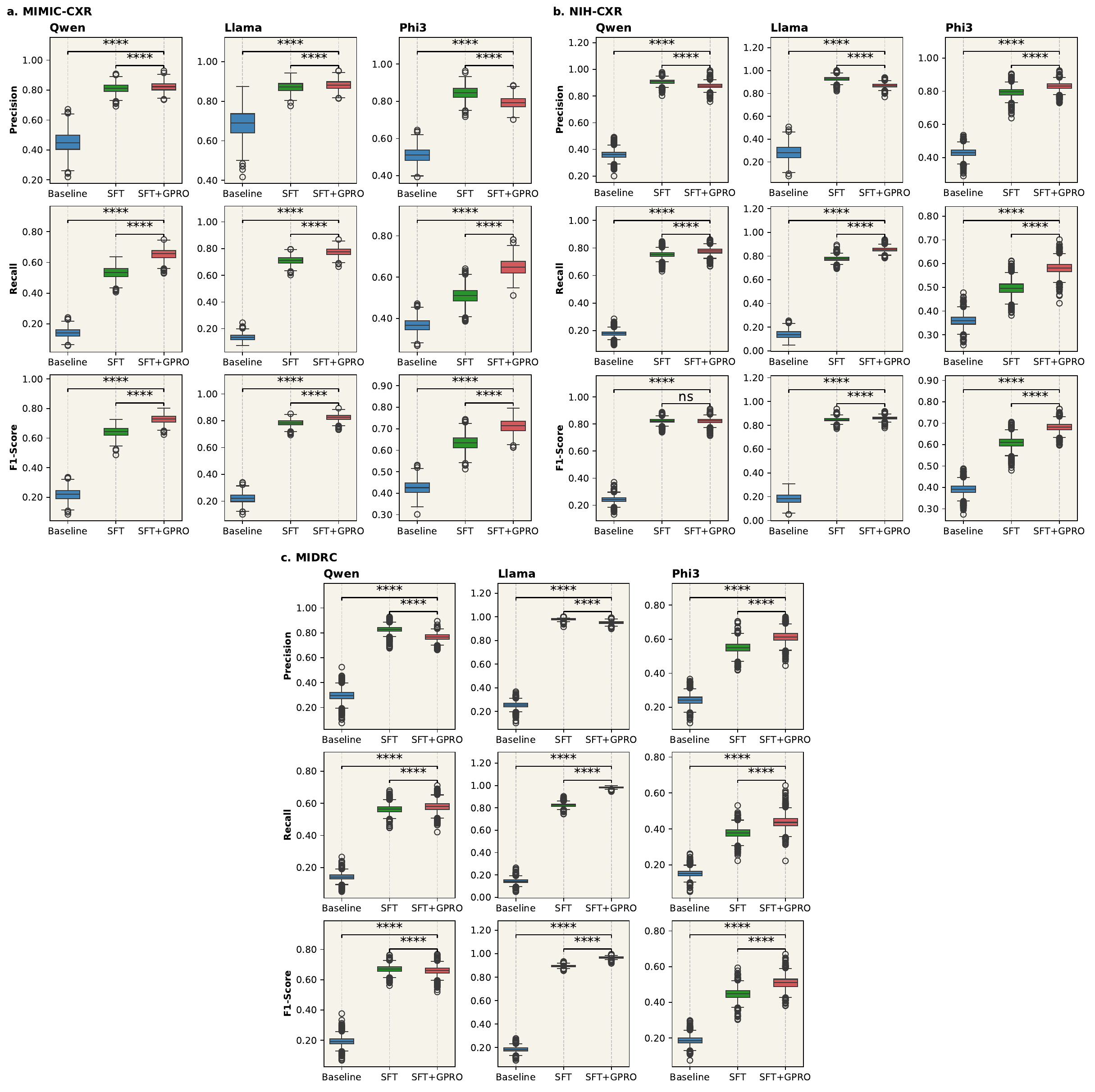}
\caption{Disease detection performance on the MIMIC-CXR, NIH-CXR, and MIDRC evaluation dataset. SFT - Supervised Fine-Tuning. SFT+GRPO - Supervised Fine-Tuning followed by Group Relative Policy Optimization. Significance levels: * - $p<0.05$; ** - $p < 0.01$; *** - $p < 0.001$; **** - $p < 0.0001$; ns - Not significant. The error bars are 95\% CIs.}
\label{fig:disease detection}
\end{figure}

Figure~\ref{fig:disease detection} (Supplementary Table~\ref{stab:variants}) demonstrates that SFT consistently outperforms the baseline across all cohorts, and the addition of GRPO yields further micro-F1 improvements in eight of nine cohorts, with the largest gains observed for Qwen on MIMIC-CXR (13.2\%), Phi3 on NIH-CXR (12.8\%), and LLaMA on MIDRC (8.0\%). Notably, for LLaMA, the final classification accuracy nearly matches that of GPT-4o, which served as the teacher model for generating all training supervision during the SFT stage.

To increase the sample size of the evaluation data, we constructed a MIMIC-CXR-Evaluation-Silver subset comprising 1,000 randomly selected reports from the remaining dataset, with labels curated by GPT-4o. As shown in  Table~\ref{stab:silver}, the results on this silver-standard subset are consistent with those obtained on the gold-standard evaluation sets. However, the performance advantage of GRPO is less pronounced, likely because the evaluation labels were generated by the same model used for SFT, rendering them susceptible to similar labeling errors. To assess the impact of model scale, we applied our framework to Qwen-7B and compared it with the Qwen-3B across all three cohorts (Supplementary Table 1). Qwen-7B consistently outperformed its smaller counterpart, yielding Micro F1 scores of 0.788 vs. 0.729 on MIMIC, 0.841 vs. 0.824 on NIH-CXR, and 0.935 vs. 0.661 on MIDRC.
\begin{table}
\caption{Prediction accuracy on MIMIC-CXR-evaluation-silver dataset.}
\label{stab:silver}
\centering
\footnotesize
\begin{tabular}{lccccccccc}
\toprule
\multirow{2}{*}{Model} & \multicolumn{3}{c}{Micro precision} & \multicolumn{3}{c}{Micro recall} & \multicolumn{3}{c}{Micro F1} \\
\cmidrule(rl){2-4}\cmidrule(rl){5-7}\cmidrule(rl){8-10}
 & Baseline & SFT & \makecell[c]{SFT\\+GRPO} & Baseline & SFT & \makecell[c]{SFT\\+GRPO} & Baseline & SFT & \makecell[c]{SFT\\+GRPO} \\
\midrule
Qwen-3B  & 0.489 & \textbf{0.787} & 0.778 & 0.389 & 0.539 & \textbf{0.648} & 0.433 & 0.640 & \textbf{0.707} \\
Llama & 0.454 & \textbf{0.775} & 0.771 & 0.436 & \textbf{0.847} & 0.839 & 0.445 & \textbf{0.809} & 0.804 \\
Phi3  & 0.405 & 0.683 & \textbf{0.708} & 0.414 & 0.548 & \textbf{0.615} & 0.409 & 0.608 & \textbf{0.658} \\
\bottomrule
\end{tabular}

\end{table}

To understand whether medicine-focused LLMs would help for this task, we applied the same strategy to MedAlpaca-7B \cite{han2023medalpaca}, an LLama-based model fine-tuned for medical question answering. As shown in Supplementary Table \ref{stab:variants}, the baseline significantly underperformed compared to general-purpose LLMs due to poor instruction following. While SFT and GRPO improved performance consistently, their results were comparable, but not superior to, those achieved by general models. This may be due to a lack of alignment between MedAlpaca’s medical QA training data and our specific task requirements.

To compare with conventional text classifiers, we fine-tuned BioClinicalBERT \cite{alsentzer2019publicly} on the same training set. Our strategy using larger base models (e.g., LLaMA) consistently achieved higher performance across the three benchmarks. Qwen's performance was comparable to or exceeded BioClinicalBERT on the MIMIC and NIH datasets, though it underperformed on MIDRC. Conversely, the Phi-3 model consistently lagged behind BioClinicalBERT across all benchmarks. Unlike LLMs, BERT-based models are limited by their ability to generate natural language reasoning to support the classifications.

To gain further insight into model performance, we reported per-disease accuracy on MIMIC-CXR (Table~\ref{stab:accuracy_per_disease}). Diseases with very few samples, such as Enlarged cardiomediastinum, Lung lesion, and Pleural other, were excluded from this analysis. Across nearly all diseases, the SFT+GRPO configuration consistently outperforms SFT alone in terms of micro-F1, primarily driven by its substantially higher recall, as discussed previously. One thing to note is that for certain conditions, such as pneumonia, the low label prevalence may reduce the statistical reliability of the result.
\begin{table}
\caption{Prediction accuracy per disease on MIMIC-CXR dataset.}
\label{stab:accuracy_per_disease}
\centering
\footnotesize
\begin{tabular}{lccccccccc}
\toprule
Disease & \multicolumn{3}{c}{Micro precision} & \multicolumn{3}{c}{Micro recall} & \multicolumn{3}{c}{Micro F1} \\
\cmidrule(rl){2-4}\cmidrule(rl){5-7}\cmidrule(rl){8-10}
 & Baseline & SFT & \makecell[c]{SFT\\+GRPO} & Baseline & SFT & \makecell[c]{SFT\\+GRPO} & Baseline & SFT & \makecell[c]{SFT\\+GRPO} \\
\hline
\rowcolor{graybg} \multicolumn{10}{c}{Llama} \\
Atelectasis      & 0.333 & 0.690 & 0.750 & 0.040 & 0.800 & 0.960 & 0.071 & 0.741 & 0.842 \\
Cardiomegaly     & 1.000 & 1.000 & 1.000 & 0.086 & 0.971 & 0.914 & 0.158 & 0.985 & 0.955 \\
Consolidation    & -    & 0.400 & 0.400 & 0.000 & 1.000 & 1.000 & -    & 0.571 & 0.571 \\
Edema            & -    & 0.944 & 0.947 & 0.000 & 0.944 & 1.000 & -    & 0.944 & 0.973 \\
Fracture         & -    & 1.000 & 1.000 & 0.000 & 0.667 & 0.667 & -    & 0.800 & 0.800 \\
Lung opacity     & 1.000 & 0.824 & 0.882 & 0.030 & 0.424 & 0.455 & 0.058 & 0.560 & 0.600 \\
Pleural effusion & 0.750 & 0.935 & 0.938 & 0.581 & 0.935 & 0.968 & 0.655 & 0.935 & 0.953 \\
Pneumonia        & 0.600 & 1.000 & 0.600 & 0.600 & 0.200 & 0.600 & 0.600 & 0.333 & 0.600 \\
Pneumothorax     & -    & 0.667 & 1.000 & 0.000 & 0.667 & 1.000 & -    & 0.667 & 1.000 \\
Support devices  & -    & 1.000 & 0.905 & 0.000 & 0.438 & 0.594 & -    & 0.609 & 0.717 \\
\rowcolor{graybg} \multicolumn{10}{c}{Qwen} \\
Atelectasis      & 0.308 & 0.737 & 0.667 & 0.160 & 0.560 & 0.640 & 0.211 & 0.636 & 0.653 \\
Cardiomegaly     & 1.000 & 0.938 & 0.941 & 0.286 & 0.857 & 0.914 & 0.445 & 0.896 & 0.927 \\
Consolidation    & 0.333 & 0.500 & 0.500 & 0.500 & 0.500 & 0.500 & 0.400 & 0.500 & 0.500 \\
Edema            & 1.000 & 0.800 & 0.778 & 0.056 & 0.667 & 0.778 & 0.106 & 0.727 & 0.778 \\
Fracture         & -    & 1.000 & 0.750 & 0.000 & 0.667 & 0.500 & -    & 0.800 & 0.600 \\
Lung opacity     & 1.000 & 0.667 & 0.917 & 0.091 & 0.182 & 0.333 & 0.167 & 0.286 & 0.489 \\
Pleural effusion & 1.000 & 0.941 & 0.913 & 0.129 & 0.516 & 0.677 & 0.229 & 0.667 & 0.777 \\
Pneumonia        & 0.400 & 0.250 & 0.500 & 0.800 & 0.200 & 0.800 & 0.533 & 0.222 & 0.615 \\
Pneumothorax     & 0.333 & 1.000 & 1.000 & 0.333 & 0.333 & 1.000 & 0.333 & 0.500 & 1.000 \\
Support devices  & -    & 0.818 & 0.955 & 0.000 & 0.563 & 0.656 & -    & 0.667 & 0.778 \\
\rowcolor{graybg} \multicolumn{10}{c}{Phi4} \\
Atelectasis      & 0.640 & 0.636 & 0.667 & 0.640 & 0.560 & 0.800 & 0.640 & 0.596 & 0.727 \\
Cardiomegaly     & 1.000 & 1.000 & 0.958 & 0.657 & 0.486 & 0.657 & 0.793 & 0.654 & 0.779 \\
Consolidation    & 0.000 & 0.250 & 0.400 & 0.000 & 0.500 & 1.000 & -    & 0.333 & 0.571 \\
Edema            & -    & 0.923 & 0.789 & 0.000 & 0.667 & 0.833 & -    & 0.774 & 0.810 \\
Fracture         & -    & 1.000 & 0.714 & 0.000 & 0.833 & 0.833 & -    & 0.909 & 0.769 \\
Lung opacity     & -    & 1.000 & 0.833 & 0.000 & 0.182 & 0.303 & -    & 0.308 & 0.444 \\
Pleural effusion & 0.800 & 0.957 & 0.871 & 0.774 & 0.710 & 0.871 & 0.787 & 0.815 & 0.871 \\
Pneumonia        & 0.238 & 0.600 & 0.333 & 1.000 & 0.600 & 0.400 & 0.384 & 0.600 & 0.363 \\
Pneumothorax     & 0.750 & 1.000 & 1.000 & 1.000 & 0.333 & 0.333 & 0.857 & 0.500 & 0.500 \\
Support devices  & -    & 0.944 & 0.952 & 0.000 & 0.531 & 0.625 & -    & 0.680 & 0.755 \\
\hline
\end{tabular}

\end{table}

\begin{figure}
\centering
\includegraphics[width=\linewidth]{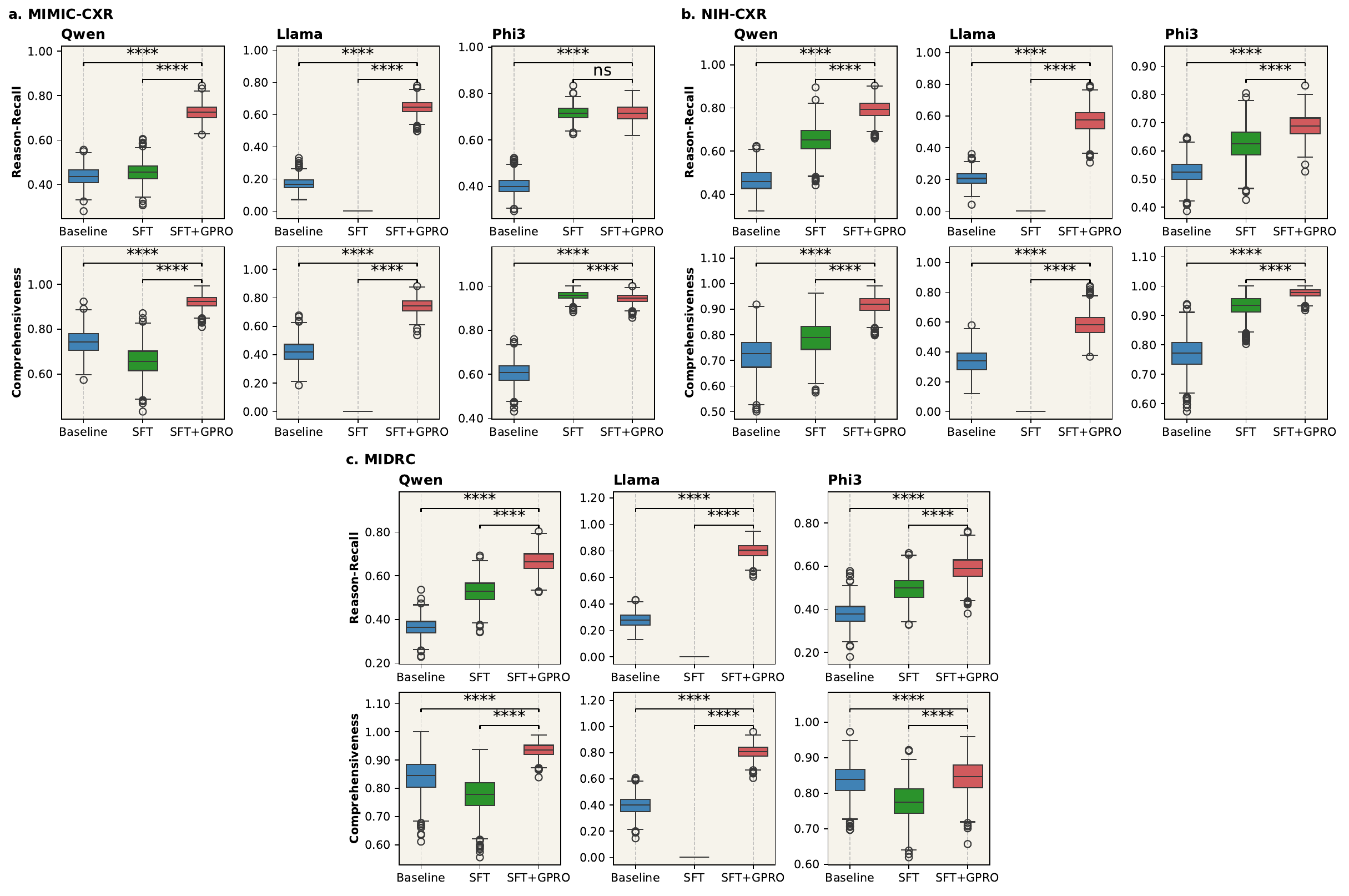}
\caption{Reasoning performance of the MIMIC-CXR, NIH-CXR, and MIDRC datasets. SFT: Supervised Fine-Tuning. SFT+GRPO: Supervised Fine-Tuning followed by Group Relative Policy Optimization. Significance levels: * - $p<0.05$; ** - $p < 0.01$; *** - $p < 0.001$; **** - $p < 0.0001$; ns - Not significant. The error bars are 95\% CIs.}
\label{fig:reasoning}
\end{figure}

Figure~\ref{fig:reasoning} (Supplementary Table~\ref{stab:reasoning}) shows the reasoning recall and comprehensiveness across the three radiologist-curated evaluation datasets.
On both Qwen and Phi3, findings collectively indicate that GRPO effectively mitigates the reasoning performance losses induced by SFT, with Qwen in particular exhibiting the largest relative gains.
After the application of GRPO, both reasoning recall and comprehensiveness show improvement in eight of the nine evaluated cohorts. 
For LLaMA, reasoning capability is completely lost after the SFT stage due to \textit{catastrophic forgetting}. The model simply reproduces the disease list without providing any insight into the thought process. 
GRPO is able to partially restore the model's reasoning ability, producing more coherent and clinically meaningful explanations.

To avoid the potential feedback loops from using the same GPT-4o as both labeler and evaluator, we reimplemented the evaluation of reasoning quality using Gemini 2.0 Flash \cite{team2024gemini}. The prompt was slightly adjusted to get meaningful reasoning due to the differences between GPT-4o and Gemini (Supplementary Prompt \ref{prompt:gemini}). Supplementary Table \ref{stab:gemini} shows that the conclusion remains consistent with that of GPT-4o, although Gemini consistently overestimates reasoning recall and reasoning comprehensiveness, as it occasionally attributes a simple mention of a disease as `covered' in reasoning, even if it is directly excerpted from the report.

In addition to the automated evaluation conducted by GPT-4o and Gemini, two board-certified radiologists (G.S., A.F.) independently reviewed the reasoning outputs. A total of fifty reports were randomly selected from the MIMIC-CXR-evaluation dataset. For each report, reasoning outputs generated by the SFT and SFT+GRPO models were randomly shuffled. The radiologists were instructed to select, for each case, the reasoning that they judged to be more accurate based on two criteria: (1) consistency between the reasoning and the corresponding report, and (2) consistency between the reasoning and the ground-truth disease list.
We found that one radiologist preferred the SFT+GRPO output in 30 cases, the SFT output in 17 cases, and reported no preference in 3 cases. The second one preferred SFT+GRPO in 18 cases, SFT in 11 cases, and had no preference in 21 cases. Among the 29 cases where both radiologists expressed a clear preference, they agreed in 20. These results provide additional evidence that GRPO improves the model's reasoning performance.

\begin{figure}
\centering
\includegraphics[width=\linewidth]{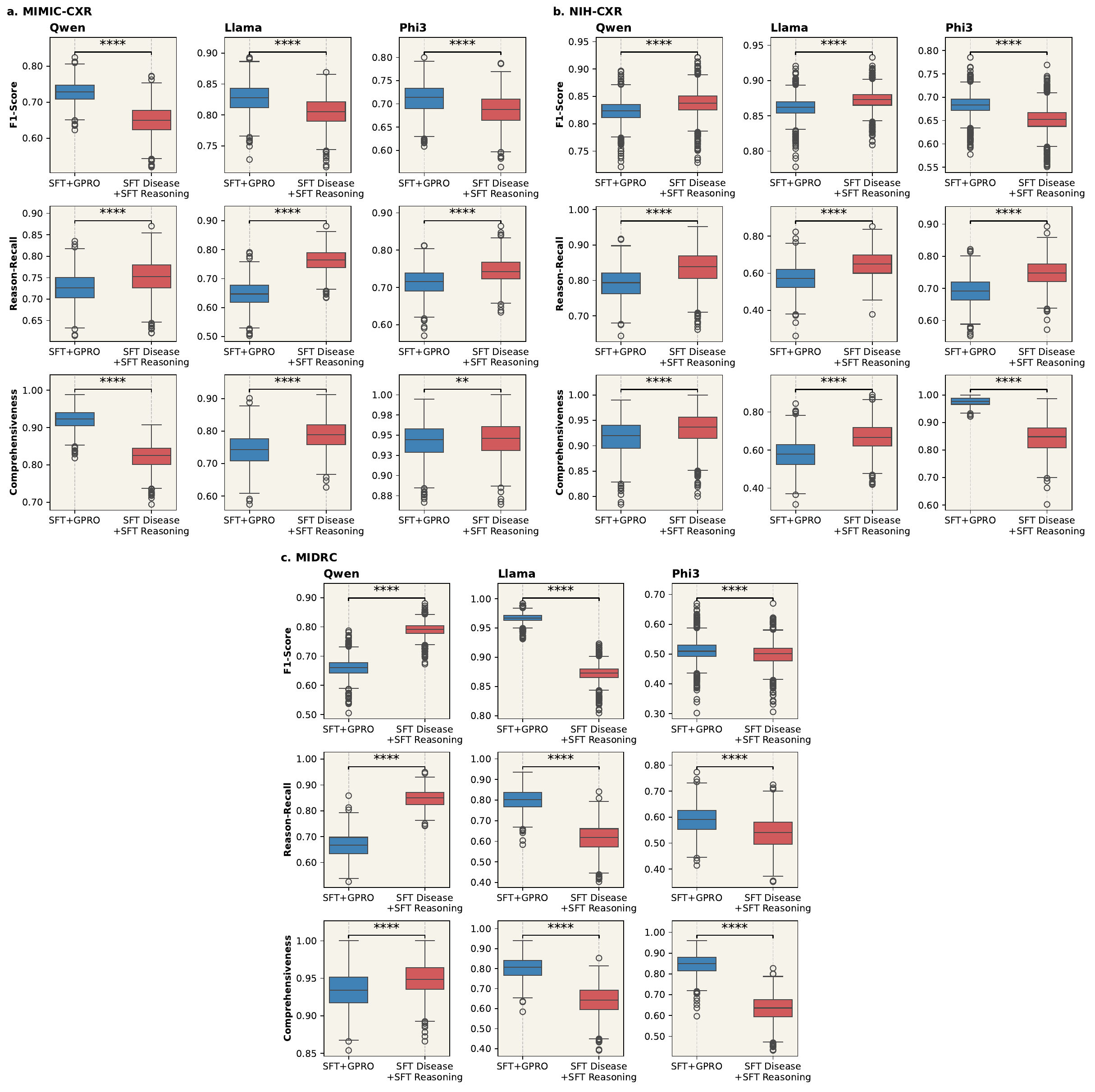}
\captionof{figure}{micro-F1 and reasoning performance of the MIMIC-CXR, NIH-CXR, and MIDRC datasets. SFT+GRPO: Supervised Fine-Tuning followed by Group Relative Policy Optimization. SFT+Reasoning: Supervised fine-tuning on disease labels, followed by additional fine-tuning on a small amount of reasoning data. Significance levels: * - $p<0.05$; ** - $p < 0.01$; *** - $p < 0.001$; **** - $p < 0.0001$; ns - Not significant. The error bars are 95\% CIs.}
\label{fig:microf1}
\end{figure}

Finally, we compared the performance of GRPO against additional fine-tuning with a limited amount of reasoning data on MIMIC-CXR (Figure~\ref{fig:microf1}; Supplementary Table~\ref{stab:comparing}). Regarding disease classification, the SFT+GRPO model generally outperforms the SFT disease+SFT reasoning one, though the degree of improvement varies across cohorts. In contrast, SFT disease+SFT reasoning yields a stronger reasoning ability than SFT+GRPO, but requires explicit reasoning supervision. Notably, both SFT+GRPO and SFT disease+SFT reasoning on a small reasoning subset lead to measurable performance gains, suggesting that strengthening reasoning capability enables lightweight LLMs to make more accurate predictions. The importance of reasoning supervision for model performance is also consistent with findings reported in other studies \cite{Fan2025-nc}.

This study has several limitations. First, the GRPO reward function only considers classification accuracy and format loss and does not explicitly supervise reasoning attributes such as reasoning length or alignment between reasoning and predicted diseases. Exploring these aspects could be valuable. Second, LLMs are used to evaluate reasoning recall and comprehensiveness, but its judgments may be imperfect and subject to ``LLM-as-a-judge'' biases \cite{gu2024survey}. Due to a lack of ground truth reasoning for validation, our assessment of reasoning capabilities primarily focused on the mention of specific disease labels. Future evaluations should incorporate clinical fidelity, such as grounding the reasoning in evidence from the original report, as well as assessing coherence and fluency. Third, GRPO is only designed to enhance the model's reasoning ability. Future work could compare or integrate GRPO with more general catastrophic forgetting mitigation techniques, such as elastic weight consolidation \cite{kirkpatrick2017overcoming} or memory replay \cite{tadros2022sleep-like}. Finally, due to the inherent nature of LLMs, obtaining calibrated probabilities or reliable uncertainty estimates for the predicted disease list remains challenging. Future work could be devoted to addressing this gap.

To summarize, this study demonstrates that the reinforcement learning algorithm GRPO enhances the reasoning ability of lightweight LLMs following supervised fine-tuning and further improves their performance in disease classification from radiology reports, all without requiring additional reasoning supervision.

\newpage

\section*{METHODS}\label{materials-and-methods}

\subsection*{Problem formulation}\label{problem-formulation}

This study aims to identify lung diseases and abnormal findings documented in radiology reports. For fine-tuning the LLM, the input consists of (1) a natural language passage corresponding to the radiology report and (2) a set of ground-truth abnormal findings (labels). This study focuses on 13 common lung diseases: Atelectasis, Cardiomegaly, Consolidation, Edema, Enlarged Cardiomediastinum, Fracture, Lung Lesion, Lung Opacity, Pleural Effusion, Pleural Other, Pneumonia, Pneumothorax, and Support Devices.
The disease distributions in Table~\ref{tab:data} reveal a heavy long-tailed pattern. In the MIMIC-CXR dataset, the most prevalent diseases are Cardiomegaly, Atelectasis, Lung opacity, Pleural effusion and Support devices. By contrast, the NIH and MIDRC datasets are primarily dominated by Lung opacity, Pleural effusion and Support devices.

The LLM is prompted to identify the diseases mentioned in the radiology report and to provide a justification for each prediction. A representative example of a valid response together with its corresponding reasoning is listed in Prompt~\ref{prompt:grpo}. The number of diseases identified may vary across reports, and an empty list is returned when no diseases are detected. Only the positive presence of diseases explicitly documented in the report is considered a valid answer.

\begin{prompt}[Prompt template for GRPO/inference/GPT reasoning generation]
\label{prompt:grpo}
\begin{lstlisting}[language=prompt]
You are given a clinical report. Your task is to identify any diseases mentioned in the report and format the answer as a list.
Instructions:
1. Only use diseases from this list:
['Atelectasis', 'Cardiomegaly', 'Consolidation', 'Edema', 'Enlarged Cardiomediastinum', 'Fracture', 'Lung Lesion', 'Lung Opacity', 'Pleural Effusion', 'Pleural Other', 'Pneumonia', 'Pneumothorax', 'Support Devices']
2. If there is no disease, return an empty list. 
3. First, think about the reasoning process in the mind and then provide the answer. The reasoning should be based on phrases or evidence from the report.
############# Example ####################
## Report:
The endotracheal tube terminates 6.9 cm above the carina.  The right subclavian line tip is at the mid SVC. The NG tube passes below the diaphragm and out of view.
## Answer:
<reasoning>Support devices is found because the report mentions: 'Endotracheal tube', 'subclavian line' and 'NG tube'.</reasoning> <answer>[Support Devices]</answer>
############# Analyze the following report:
{report to analyze}
\end{lstlisting}
\end{prompt}

\subsection*{Study design and data collection}\label{study-design-and-data-collection}

In this retrospective study, we utilized three datasets of de-identified, structured radiology reports (Table~\ref{tab:data}). Each report included a \textit{Findings} section and an \textit{Impressions} section, with additional sections such as \textit{History}, \textit{Comparison}, and \textit{Exam type} available for some reports. For all evaluation datasets, radiologists were instructed to re-examine the entire report and list all disease labels that they could identify. 

\subparagraph{MIMIC-CXR} 

MIMIC-CXR is a large, publicly available collection of chest X-ray examinations from Beth Israel Deaconess Medical Center, comprising imaging studies conducted between 2011 and 2016 \cite{Johnson2019-eb}. For this study, four distinct subsets were curated from the original MIMIC-CXR dataset. 
\begin{enumerate}[nosep]
\item \textbf{Training}. This subset consists of 2,000 randomly sampled reports. Labels for the 13 targeted diseases were automatically extracted using GPT-4o (gpt-4o-2024-05-13) through prompting (Prompt~\ref{prompt:generation}). This dataset was used exclusively for supervised fine-tuning. Because these labels were generated for training purposes only, no formal quality assessment was performed. This relatively small size reflects the scarcity of labeled data in real-world clinical settings.
\item \textbf{Reasoning}. This subset includes 200 reports randomly sampled from the training set. In addition to disease labels, GPT-4o was prompted to generate explicit reasoning for each disease label. 
\item \textbf{Evaluation}. This subset includes 100 reports randomly selected from the remaining MIMIC-CXR dataset. All reports were manually annotated by a board-certified radiologist (HO) and a domain expert (YW). Inter-rater agreement, measured using Cohen's Kappa coefficient \cite{Waisberg2023-fu}, was 0.88. Discrepancies primarily stemmed from uncertain disease labels or occasional omissions and were resolved through consensus to ensure the quality of the final ground truth annotations used for model evaluation. 
\item \textbf{Evaluation-Silver}. This supplementary subset consists of 1,000 randomly selected reports from the remaining dataset, with labels curated by GPT-4o. It was created to augment the limited sample size of the gold-standard evaluation set.
\end{enumerate}

\subparagraph{NIH-CXR} 

The second dataset consists of 50 reports randomly selected from the National Institutes of Health (NIH) chest radiographic dataset \cite{Wang2017-lm}, with each report dictated by one attending physician and three radiology residents. This dataset has been previously reported \cite{Zhou2024-ib}.

\subparagraph{MIRDC} 

The third dataset consists of 50 reports randomly selected from the Medical Imaging and Data Resource Center (MIDRC, \url{https://www.midrc.org}). This dataset has also been previously reported \cite{Zhou2024-ib}.

The Institutional Review Board waived the requirement to obtain written informed consent from all patients because the data sets used in this study were publicly available.

\begin{prompt}[Prompt template for SFT/GPT result only data generation]
\label{prompt:generation}
\begin{lstlisting}[language=prompt]
You are given a clinical report. Your task is to identify any diseases mentioned in the report and format the answer as a list.
Instructions:
1. Only use diseases from this list: ['Atelectasis', 'Cardiomegaly', 'Consolidation', 'Edema', 'Enlarged Cardiomediastinum', 'Fracture', 'Lung Lesion', 'Lung Opacity', 'Pleural Effusion', 'Pleural Other', 'Pneumonia', 'Pneumothorax', 'Support Devices']
2. If there is no disease, return an empty list. 
############# Analyze the following report:
{note}
\end{lstlisting}
\end{prompt}

\subsection*{Development of the proposed model}\label{development-of-the-proposed-model}

We propose a \textbf{two-stage pipeline} designed to enhance both the model's disease classification performance and its ability to generate reasoning that justifies its predictions (Figure~\ref{fig:workflow}).

\paragraph{Stage 1: Supervised Fine-Tuning (SFT) on disease labels.} 

\begin{sloppypar}
In the first stage, SFT is performed using the reports and corresponding disease labels from the \textbf{MIMIC-CXR-Training} dataset. The instruction prompt used during fine-tuning closely resembles Prompt~\ref{prompt:generation}, with the exception that no examples are included. Labels follow the format: \texttt{<reasoning></reasoning><answer>...</answer>}, with the reasoning section intentionally left empty, reflecting the absence of explicit reasoning supervision. No parameter-efficient fine-tuning (PEFT) techniques, such as LORA \cite{Hu2021-qq}, are applied at this stage.
\end{sloppypar}

\paragraph{Stage 2: Reinforcement learning with GRPO.} 

In the second stage, GRPO is applied to the same dataset of 2,000 reports used in Stage 1, but with a modified loss function. Following the approach described in Guo et al. \cite{DeepSeek-AI2025-pz}, the reward function verifies both \textbf{classification accuracy} and \textbf{formatting correctness}.

Classification accuracy is calculated as the average of precision and recall. If the model makes no predictions, both precision and recall are set to 1 in the absence of ground-truth labels. If ground truth labels exist, both are set to 0. 

\begin{sloppypar}
Formatting correctness is designed to encourage adherence to the required structure: \texttt{<reasoning>...</reasoning><answer>...</answer>}. Unlike previous work, the formatting score is set to zero if the reasoning section in the response is empty or simply repeats the predicted disease list, thus promoting meaningful reasoning output.
\end{sloppypar}

To demonstrate the desired format, the prompt used during GRPO training is identical to the one used during inference, which differs from the SFT prompt (Prompt~\ref{prompt:generation}) by including an explicit example that demonstrates how the reasoning section should be completed. The final reward function is defined as a weighted sum of the two components.
\[
\text{Reward} = 0.8 \times \text{accuracy reward} + 0.2 \times \text{formatting reward}
\]

\subsection*{Inference with majority voting and reasoning summarization}\label{inference-with-majority-voting-and-reasoning-summarization}

For each radiology report, five independent classifications and their corresponding reasoning outputs are generated. The final disease classification is determined using a \textbf{majority voting} procedure. To consolidate the reasoning outputs into a single coherent explanation, the original lightweight LLM is prompted to summarize the original set of individual reasoning outputs to justify the final ensemble disease list (Prompt~\ref{prompt:summarize}). It is worth noting that we did not optimize the ensemble size (fixed at five in this study). Future research could investigate tuning this hyperparameter to achieve an optimal balance between inference cost and performance.

This ensembling strategy provides two key advantages. First, classification diversity increases significantly following GRPO training, and majority voting leverages this diversity to improve overall classification performance. Second, generating multiple inferences reduces the likelihood of omitting critical information. The final summarization step further enhances the quality of reasoning by organizing, verifying, and deduplicating information across outputs. For all inferences, the temperature is set to 0.1 and the top\_p parameter is set to 1.

\begin{prompt}[Prompt to summarize reasonings from the five inferences]
\label{prompt:summarize}
\begin{lstlisting}[language=prompt]
You are a radiology specialist. You are provided with five reasons for the radiology report prediction of the same report. You are also given a final disease list, which is based on the majority voting of the five predictions. Your task is to summarize the five reasons and get an overall reasoning report.
### Instructions
1. The reasoning should be focused on explaining the final disease list. So you don't need to include all points from the provided reasoning.
2. The reasoning should mostly be finding evidence from the original report.
3. The provided reasonings may not be of high quality, so you need to paraphrase and summarize them.
4. Summarize the report naturally, so as not to show that you are summarizing. Output is a natural reasoning report.
5. DO NOT make up any evidence. ONLY use evidence from the provided reasonings. If all reasonings are empty, return empty output as well.
### Input:
    ##  reason0: {reason_0}
    ##  reason1: {reason_1}
    ##  reason2: {reason_2}
    ##  reason3: {reason_3}
    ##  reason4: {reason_4}
    ## disease_list: {disease_ensemble}
### Your summarized reasoning:
\end{lstlisting}
\end{prompt}

\subsection*{Experiment settings}\label{experiment-settings}

We evaluated three lightweight LLMs capable of local deployment: LLaMA 3.1-8B-Instruct (LLaMa) \cite{Grattafiori2024-rd}, Qwen 2.5-3B-Instruct (Qwen) \cite{Qwen2024-mg}, and Phi-3 Min-128K-Instruct (Phi3) \cite{Abdin2024-uo}. Medical focused LLM MedAlpaca-7B \cite{han2023medalpaca} and conventional medical text classifical BioClinicalBERT \cite{alsentzer2019publicly} was also used for ablation studies. Stage 1 consisted of SFT on disease labels using 2,000 reports. Stage 2 applied GRPO training on the same dataset. To offer a more comprehensive assessment, we also explored additional configurations in which GRPO was replaced by additional SFT on 200 reports (MIMIC-CXR-Reasoning).

In total, for each lightweight LLM, we evaluated five configurations: 
\begin{enumerate}[nosep]
\item \textbf{Base}: the pre-trained model without fine-tuning;
\item \textbf{SFT}: fine-tuned on 2,000 reports using disease labels;
\item \textbf{SFT+GRPO}: SFT on 2,000 reports followed by GRPO (our proposed pipeline);
\item \textbf{SFT reasoning}: SFT on 200 reports with explicit reasoning supervision;
\item \textbf{SFT disease + SFT reasoning}: SFT on 2,000 reports with disease labels, followed by SFT on 200 reports with explicit reasoning supervision. 
\end{enumerate}

In addition, GPT-4o was included as a reference model for comparison.

\subsection*{Evaluation metrics}\label{evaluation-metrics}

For the disease classification task, we evaluated performance using micro-averaged precision, recall, and F1 score. To quantify uncertainty, we generated 1,000 bootstrap samples from the test set and reported 95\% confidence intervals. For each bootstrap iteration, we sampled \(n\) reports with replacement from the test set of size \(n\).

To assess the quality of generated reasoning, we propose two metrics. Let \(L\) be the ground-truth disease list and \(G\) the disease list predicted by the LLM. For a reasoning response \(y\), let \(\hat{L}_{y}\) represent the set of disease labels mentioned in \(y\). \textbf{Reasoning recall} is defined as \(R(y) = \frac{1}{|L|}\sum_{l \in L}\mathbbm{1}[l \in \hat{L}_{y}]\). This measures the proportion of ground-truth diseases referenced in the reasoning. \textbf{Reasoning comprehensiveness} is defined as \(C(y) = \frac{1}{|G|}\sum_{g \in G}\mathbbm{1}[ g \in \hat{L}_{y}]\). This quantifies the proportion of predicted diseases that are supported by explicit reasoning.

Note that we did not evaluate reasoning \textbf{precision}, defined as the proportion of target diseases in the reasoning that match the true labels (\(\frac{1}{|\hat{L}_{y}|}\sum_{l \in \hat{L}_{y}} \mathbbm{1}[l \in L]\)), because target diseases are often expressed negatively (e.g., \textit{no pneumonia}) and should not be included in the ground-true disease list. As such, precision is not an appropriate metric in this context.

We used GPT-4o to calculate both reasoning metrics with three inputs: the original radiology report, the predicted disease list, and the generated reasoning. GPT-4o was then prompted to evaluate the reasoning in a structured output format (Prompt~\ref{prompt:eval} and Prompt~\ref{prompt:schema}). Specifically, it first identified key phrases in the reasoning. Then, it determined whether each phrase was supported by the content of the original report and the disease it intended to explain. These metrics and the use of GPT-4o as the evaluator have been proposed in prior studies \cite{Fan2025-nc}.

\begin{prompt}[GPT evaluation reasoning script]
\label{prompt:eval}
\begin{lstlisting}[language=prompt]
There is an AI assistant that tends to extract diseases from radiology reports. You are given the report, the reasoning given by the assistant and the result given by the assistant. Your task is to evaluate whether the AI assistant is doing a correct job.
********* Instructions:
1. The output is a list of formatted structures, each element in the list has the components of: 'phrase', 'whether supported by report', 'target diseases', 'whether lead to the answer'
2. For the 'phrase' component, your task is to divide the reasoning into semantically independent parts. Each part will lead to one structured element returned, and the part will be the 'phrase' component.
3. 'whether_supported_by_report' is a boolean to evaluate whether the 'phrase' is supported by the report.
4. 'target_diseases': which diseases are targeted by this phrase. Here all diseases come in this list: ['Atelectasis', 'Cardiomegaly', 'Consolidation', 'Edema', 'Enlarged Cardiomediastinum', 'Fracture', 'Lung Lesion', 'Lung Opacity', 'Pleural Effusion', 'Pleural Other', 'Pneumonia', 'Pneumothorax', 'Support Devices'] Each phrase can target to a several diseases or even no disease. So 'target diseases' are a list of candidate disease this 'phrase' is targeting for
5. If there is no reasoning or the reasoning is clearly just repeating a list or not making sense, return empty list
********** Example
## Report
FINDINGS: A cluster of heterogeneous opacities in the right lower lung has continued to grow since ___. Otherwise, the lungs are clear. Moderate cardiomegaly, including severe left atrial enlargement, is chronic; there is no pulmonary vascular congestion or edema. The thoracic aorta is heavily calcified.  There may be a new small right pleural effusion or pneumothorax.

IMPRESSION: Slowly progressive chronic right pneumonia, could be exogenous lipoid pneumonia, but tuberculosis is in the differential.  CT scanning is recommended.  Nurse ___ and I discussed the findings and their clinical significance by telephone at the time of dictation.

## Reasoning
According to the report, there is a cluster of heterogeneous opacities in the right lower lung. A pneumonia has developed.  There are bilateral pleural effusions, one on the right and one on the left. Additionally, there may be a small pneumothorax at the right lung base.  Supportive devices are mentioned, like a chest tube, but not specified.

## Result
['Pneumonia', 'Pleural Effusion', 'Pneumothorax', 'Support Devices']   
        
###### Your output should be:
\end{lstlisting}
\begin{lstlisting}[language=json]
[{{
    'phrase': 'According to the report, there is a cluster of heterogeneous opacities in the right lower lung.',
    'whether_supported_by_report': True,
    'target_diseases': ['Lung Opacity'],
}},
{{
    'phrase': 'A pneumonia has developed',
    'whether_supported_by_report': True,  (The report mentions 'Slowly progressive chronic right pneumonia')
    'target_diseases': ['Pneumonia'],
}},
{{
    'phrase': 'There are bilateral pleural effusions, one on the right and one on the left.',
    'whether_supported_by_report': False,  (The report mentions it is on the right)
    'target_diseases': ['Pleural effusion'],
}},
{{
    'phrase': 'Additionally, there may be a small pneumothorax at the right lung base.', 
    'whether_supported_by_report': True,  (Although the report mentions on the right, not right base, but it is very close, so mark as True)
    'target_diseases': ['Pneumothorax'],
}},
{{
    'phrase': 'Supportive devices are mentioned, like a chest tube, but not specified.',
    'whether_supported_by_report': False,  
    'target_diseases': ['Support Devices'],
}}]
\end{lstlisting}
\begin{lstlisting}[language=prompt]
********** Here is your input
    ## Report
        {note}
    ## Reasoning
        {reasoning}
    ## Result
        {result}
********** Give your output
\end{lstlisting}
\end{prompt}

\begin{prompt}[GPT evaluation output schema]
\label{prompt:schema}
\begin{lstlisting}[language=json]
{
    "type": "json_schema",
    "json_schema": {
      "name": "medical_phrase_extraction",
      "strict": True,
      "schema": {
        "type": "object",
        "properties": {
          "results": {
            "type": "array",
            "items": {
              "type": "object",
              "properties": {
                "phrase": {
                  "type": "string",
                  "description": "Extracted phrase relevant to medical findings."
                },
                "whether_supported_by_report": {
                  "type": "boolean",
                  "description": "Indicates if the phrase is supported by the report content."
                },
                "target_disease": {
                  "type": "string",
                  "description": "The disease or condition associated with the phrase."
                }
              },
              "required": ["phrase", "whether_supported_by_report", "target_disease"],
              "additionalProperties": False
            }
          }
        },
        "required": ["results"],
        "additionalProperties": False
      }
    }
}
\end{lstlisting}
\end{prompt}

\subsection*{Implementation details}\label{implementation-details}

Both SFT and GRPO were implemented using the Transformer Reinforcement Learning package \cite{von-Werra2020-ca}. All SFT training was conducted on a single A100 GPU without employing PEFT techniques.

For GRPO training, the Qwen and Phi3 models were trained on two A100 GPUs, with Flash Attention enabled \cite{Dao2022-wv}. The LLaMA model and MedAlpaca used Low-Rank Adaptation (LoRA)  \cite{Hu2021-qq}. LoRA was applied specifically to the attention modules (k-proj, q-proj, and v-proj matrices) and the output module. The LoRA hyperparameters are set as follows: a rank of 16, a scaling factor (LoRA alpha) of 32, and a dropout rate of 0.05. The learning rate was set to 1e-6, with the AdamW optimizer and bfloat16 (bf16) precision enabled.

All settings were trained for one epoch, with the exception of BioClinicalBERT, which was trained for 10 epochs based on validation error.

\section*{Data availability}\label{data_availability}

MIMIC-CXR, NIH-CXR, and MIDRC used in this study are publicly available.

\section*{Code availability}\label{code_availability}
The data and code underlying this article are available at \url{https://github.com/bionlplab/radiology-disease-classification}.

\section*{Acknowledgments}\label{fundings}

This work was supported by the National Institute of Biomedical Imaging and Bioengineering under grant number 75N920202D00021 (to YP, GS, and AF) and NSF CAREER Award under grant number 2145640 (to YP). The funder had no role in the study design; data collection, analysis, or interpretation; manuscript writing; or the decision to submit for publication. The content is solely the responsibility of the authors and does not necessarily represent the official views of the funders.

\section*{Author contributions}\label{author_contributions}

Study concepts/study design, YW, YP; manuscript drafting or manuscript revision for important intellectual content, all authors; approval of final version of the submitted manuscript, all authors; agrees to ensure any questions related to the work are appropriately resolved, all authors; literature research, YW, YL; experimental studies, YW, YL; data interpretation and statistical analysis, YW, YL, YP; data annotation, AF, GS; and manuscript editing, all authors.

\section*{Competing Interests}\label{conflicts-of-interest}

No conflicting relationships exist for any author.

\bibliographystyle{unsrtnat}
\bibliography{ref}

\begin{thebibliography}{27}
\providecommand{\natexlab}[1]{#1}
\providecommand{\url}[1]{\texttt{#1}}
\expandafter\ifx\csname urlstyle\endcsname\relax
  \providecommand{\doi}[1]{doi: #1}\else
  \providecommand{\doi}{doi: \begingroup \urlstyle{rm}\Url}\fi

\bibitem[Jain et~al.(2021)Jain, Agrawal, Saporta, Truong, Duong, Bui, Chambon, Zhang, Lungren, Ng, Langlotz, and Rajpurkar]{jain2021radgraph}
Saahil Jain, Ashwin Agrawal, Adriel Saporta, Steven Truong, Du~Nguyen Duong, Tan Bui, Pierre Chambon, Yuhao Zhang, Matthew~P Lungren, Andrew~Y Ng, Curtis Langlotz, and Pranav Rajpurkar.
\newblock {RadGraph}: Extracting clinical entities and relations from radiology reports.
\newblock In \emph{Thirty-fifth Conference on Neural Information Processing Systems Datasets and Benchmarks Track (Round 1)}, pages 1--12, 8~June 2021.

\bibitem[Chambon et~al.(2024)Chambon, Delbrouck, Sounack, Huang, Chen, Varma, Truong, Chuong, and Langlotz]{chambon2024chexpert}
Pierre Chambon, Jean-Benoit Delbrouck, Thomas Sounack, Shih-Cheng Huang, Zhihong Chen, Maya Varma, Steven Q~H Truong, Chu~The Chuong, and Curtis~P Langlotz.
\newblock {CheXpert} plus: Augmenting a large chest {X}-ray dataset with text radiology reports, patient demographics and additional image formats.
\newblock \emph{arXiv [cs.CL]}, 29~May 2024.

\bibitem[Tran et~al.(2024)Tran, Yang, Yao, and Yu]{Tran2024-wm}
Hieu Tran, Zhichao Yang, Zonghai Yao, and Hong Yu.
\newblock {BioInstruct}: instruction tuning of large language models for biomedical natural language processing.
\newblock \emph{J. Am. Med. Inform. Assoc.}, 31\penalty0 (9):\penalty0 1821--1832, 1~September 2024.
\newblock ISSN 1067-5027,1527-974X.
\newblock \doi{10.1093/jamia/ocae122}.

\bibitem[Garg et~al.(2025)Garg, Raza, Rayana, Liu, and Sohn]{Garg2025-hs}
Muskan Garg, Shaina Raza, Shebuti Rayana, Xingyi Liu, and Sunghwan Sohn.
\newblock The rise of small language models in healthcare: A comprehensive survey.
\newblock \emph{arXiv [cs.CL]}, 23~April 2025.

\bibitem[Ben~Shoham and Rappoport(2024)]{Ben-Shoham2024-ha}
Ofir Ben~Shoham and Nadav Rappoport.
\newblock {CPLLM}: Clinical prediction with large language models.
\newblock \emph{PLOS Digit. Health}, 3\penalty0 (12):\penalty0 e0000680, December 2024.
\newblock ISSN 2767-3170.
\newblock \doi{10.1371/journal.pdig.0000680}.

\bibitem[Jiang et~al.(2024)Jiang, Irvin, Ng, and Zou]{Jiang2024-ze}
Yixing Jiang, Jeremy~A Irvin, Andrew~Y Ng, and James Zou.
\newblock {VetLLM}: Large language model for predicting diagnosis from veterinary notes.
\newblock \emph{Pac. Symp. Biocomput.}, 29:\penalty0 120--133, 2024.
\newblock ISSN 2335-6928,2335-6936.

\bibitem[Wei et~al.(2024)Wei, Wang, Ong, Zhou, Flanders, Shih, and Peng]{Wei2024-dm}
Yishu Wei, Xindi Wang, Hanley Ong, Yiliang Zhou, Adam Flanders, George Shih, and Yifan Peng.
\newblock Enhancing disease detection in radiology reports through fine-tuning lightweight {LLM} on weak labels.
\newblock \emph{arXiv [cs.AI]}, 24~September 2024.

\bibitem[Guo et~al.(2025)Guo, Yang, Zhang, Song, Wang, Zhu, Xu, Zhang, Ma, Bi, Zhang, Yu, Wu, Wu, Gou, Shao, Li, Gao, Liu, Xue, Wang, Wu, Feng, Lu, Zhao, Deng, Ruan, Dai, Chen, Ji, Li, Lin, Dai, Luo, Hao, Chen, Li, Zhang, Xu, Ding, Gao, Qu, Li, Guo, Li, Chen, Yuan, Tu, Qiu, Li, Cai, Ni, Liang, Chen, Dong, Hu, You, Gao, Guan, Huang, Yu, Wang, Zhang, Zhao, Wang, Zhang, Xu, Xia, Zhang, Zhang, Tang, Zhou, Li, Wang, Li, Tian, Huang, Zhang, Wang, Chen, Du, Ge, Zhang, Pan, Wang, Chen, Jin, Chen, Lu, Zhou, Chen, Ye, Wang, Yu, Zhou, Pan, Li, Zhou, Wu, Yun, Pei, Sun, Wang, Zeng, Liu, Liang, Gao, Yu, Zhang, Xiao, An, Liu, Wang, Chen, Nie, Cheng, Liu, Xie, Liu, Yang, Li, Su, Lin, Li, Jin, Shen, Chen, Sun, Wang, Song, Zhou, Wang, Shan, Li, Wang, Wei, Zhang, Xu, Li, Zhao, Sun, Wang, Yu, Zhang, Shi, Xiong, He, Piao, Wang, Tan, Ma, Liu, Guo, Ou, Wang, Gong, Zou, He, Xiong, Luo, You, Liu, Zhou, Zhu, Huang, Li, Zheng, Zhu, Ma, Tang, Zha, Yan, Ren, Ren, Sha, Fu, Xu, Xie, Zhang, Hao, Ma, Yan, Wu, Gu, Zhu, Liu, Li, Xie, Song,
  Pan, Huang, Xu, Zhang, and Zhang]{DeepSeek-AI2025-pz}
Daya Guo, Dejian Yang, Haowei Zhang, Junxiao Song, Peiyi Wang, Qihao Zhu, Runxin Xu, Ruoyu Zhang, Shirong Ma, Xiao Bi, Xiaokang Zhang, Xingkai Yu, Yu~Wu, Z~F Wu, Zhibin Gou, Zhihong Shao, Zhuoshu Li, Ziyi Gao, Aixin Liu, Bing Xue, Bingxuan Wang, Bochao Wu, Bei Feng, Chengda Lu, Chenggang Zhao, Chengqi Deng, Chong Ruan, Damai Dai, Deli Chen, Dongjie Ji, Erhang Li, Fangyun Lin, Fucong Dai, Fuli Luo, Guangbo Hao, Guanting Chen, Guowei Li, H~Zhang, Hanwei Xu, Honghui Ding, Huazuo Gao, Hui Qu, Hui Li, Jianzhong Guo, Jiashi Li, Jingchang Chen, Jingyang Yuan, Jinhao Tu, Junjie Qiu, Junlong Li, J~L Cai, Jiaqi Ni, Jian Liang, Jin Chen, Kai Dong, Kai Hu, Kaichao You, Kaige Gao, Kang Guan, Kexin Huang, Kuai Yu, Lean Wang, Lecong Zhang, Liang Zhao, Litong Wang, Liyue Zhang, Lei Xu, Leyi Xia, Mingchuan Zhang, Minghua Zhang, Minghui Tang, Mingxu Zhou, Meng Li, Miaojun Wang, Mingming Li, Ning Tian, Panpan Huang, Peng Zhang, Qiancheng Wang, Qinyu Chen, Qiushi Du, Ruiqi Ge, Ruisong Zhang, Ruizhe Pan, Runji Wang, R~J Chen, R~L
  Jin, Ruyi Chen, Shanghao Lu, Shangyan Zhou, Shanhuang Chen, Shengfeng Ye, Shiyu Wang, Shuiping Yu, Shunfeng Zhou, Shuting Pan, S~S Li, Shuang Zhou, Shaoqing Wu, Tao Yun, Tian Pei, Tianyu Sun, T~Wang, Wangding Zeng, Wen Liu, Wenfeng Liang, Wenjun Gao, Wenqin Yu, Wentao Zhang, W~L Xiao, Wei An, Xiaodong Liu, Xiaohan Wang, Xiaokang Chen, Xiaotao Nie, Xin Cheng, Xin Liu, Xin Xie, Xingchao Liu, Xinyu Yang, Xinyuan Li, Xuecheng Su, Xuheng Lin, X~Q Li, Xiangyue Jin, Xiaojin Shen, Xiaosha Chen, Xiaowen Sun, Xiaoxiang Wang, Xinnan Song, Xinyi Zhou, Xianzu Wang, Xinxia Shan, Y~K Li, Y~Q Wang, Y~X Wei, Yang Zhang, Yanhong Xu, Yao Li, Yao Zhao, Yaofeng Sun, Yaohui Wang, Yi~Yu, Yichao Zhang, Yifan Shi, Yiliang Xiong, Ying He, Yishi Piao, Yisong Wang, Yixuan Tan, Yiyang Ma, Yiyuan Liu, Yongqiang Guo, Yuan Ou, Yuduan Wang, Yue Gong, Yuheng Zou, Yujia He, Yunfan Xiong, Yuxiang Luo, Yuxiang You, Yuxuan Liu, Yuyang Zhou, Y~X Zhu, Yanping Huang, Yaohui Li, Yi~Zheng, Yuchen Zhu, Yunxian Ma, Ying Tang, Yukun Zha, Yuting Yan,
  Z~Z Ren, Zehui Ren, Zhangli Sha, Zhe Fu, Zhean Xu, Zhenda Xie, Zhengyan Zhang, Zhewen Hao, Zhicheng Ma, Zhigang Yan, Zhiyu Wu, Zihui Gu, Zijia Zhu, Zijun Liu, Zilin Li, Ziwei Xie, Ziyang Song, Zizheng Pan, Zhen Huang, Zhipeng Xu, Zhongyu Zhang, and Zhen Zhang.
\newblock {DeepSeek}-{R1} incentivizes reasoning in {LLMs} through reinforcement learning.
\newblock \emph{Nature}, 645\penalty0 (8081):\penalty0 633--638, 18~September 2025.
\newblock ISSN 1476-4687,0028-0836.
\newblock \doi{10.1038/s41586-025-09422-z}.

\bibitem[Setlur et~al.(2024)Setlur, Garg, Geng, Garg, Smith, and Kumar]{Setlur2024-zp}
Amrith~Rajagopal Setlur, Saurabh Garg, Xinyang Geng, Naman Garg, Virginia Smith, and Aviral Kumar.
\newblock {RL} on incorrect synthetic data scales the efficiency of {LLM} math reasoning by eight-fold.
\newblock \emph{Neural Inf Process Syst}, abs/2406.14532:\penalty0 43000--43031, 20~June 2024.

\bibitem[Havrilla et~al.(2024)Havrilla, Du, Raparthy, Nalmpantis, Dwivedi-Yu, Zhuravinskyi, Hambro, Sukhbaatar, and Raileanu]{Havrilla2024-dn}
Alex Havrilla, Yuqing Du, Sharath~Chandra Raparthy, Christoforos Nalmpantis, Jane Dwivedi-Yu, Maksym Zhuravinskyi, Eric Hambro, Sainbayar Sukhbaatar, and Roberta Raileanu.
\newblock Teaching large language models to reason with reinforcement learning.
\newblock \emph{arXiv [cs.LG]}, 7~March 2024.

\bibitem[Wang et~al.(2017)Wang, Peng, Lu, Lu, Bagheri, and Summers]{Wang2017-lm}
Xiaosong Wang, Yifan Peng, Le~Lu, Zhiyong Lu, Mohammadhadi Bagheri, and Ronald~M Summers.
\newblock {ChestX}-{Ray8}: Hospital-scale chest {X}-ray database and benchmarks on weakly-supervised classification and localization of common thorax diseases.
\newblock In \emph{2017 IEEE Conference on Computer Vision and Pattern Recognition (CVPR)}, pages 3462--3471, July 2017.
\newblock \doi{10.1109/CVPR.2017.369}.

\bibitem[Zhou et~al.(2024)Zhou, Ong, Kennedy, Wu, Kazam, Hentel, Flanders, Shih, and Peng]{Zhou2024-ib}
Yiliang Zhou, Hanley Ong, Patrick Kennedy, Carol~C Wu, Jacob Kazam, Keith Hentel, Adam Flanders, George Shih, and Yifan Peng.
\newblock Evaluating {GPT}-{V4} ({GPT}-4 with vision) on detection of radiologic findings on chest radiographs.
\newblock \emph{Radiology}, 311\penalty0 (2):\penalty0 e233270, May 2024.
\newblock ISSN 0033-8419,1527-1315.
\newblock \doi{10.1148/radiol.233270}.

\bibitem[Grattafiori et~al.(2024)Grattafiori, Dubey, Jauhri, Pandey, Kadian, Al-Dahle, Letman, Mathur, Schelten, Vaughan, Yang, Fan, Goyal, Hartshorn, Yang, Mitra, Sravankumar, Korenev, Hinsvark, Rao, Zhang, Rodriguez, Gregerson, Spataru, Roziere, Biron, Tang, Chern, Caucheteux, Nayak, Bi, Marra, McConnell, Keller, Touret, Wu, Wong, Ferrer, Nikolaidis, Allonsius, Song, Pintz, Livshits, Wyatt, Esiobu, Choudhary, Mahajan, Garcia-Olano, Perino, Hupkes, Lakomkin, AlBadawy, Lobanova, Dinan, Smith, Radenovic, Guzm\'{a}n, Zhang, Synnaeve, Lee, Anderson, Thattai, Nail, Mialon, Pang, Cucurell, Nguyen, Korevaar, Xu, Touvron, Zarov, Ibarra, Kloumann, Misra, Evtimov, Zhang, Copet, Lee, Geffert, Vranes, Park, Mahadeokar, Shah, van~der Linde, Billock, Hong, Lee, Fu, Chi, Huang, Liu, Wang, Yu, Bitton, Spisak, Park, Rocca, Johnstun, Saxe, Jia, Alwala, Prasad, Upasani, Plawiak, Li, Heafield, Stone, El-Arini, Iyer, Malik, Chiu, Bhalla, Lakhotia, Rantala-Yeary, van~der Maaten, Chen, Tan, Jenkins, Martin, Madaan, Malo, Blecher,
  Landzaat, de~Oliveira, Muzzi, Pasupuleti, Singh, Paluri, Kardas, Tsimpoukelli, Oldham, Rita, Pavlova, Kambadur, Lewis, Si, Singh, Hassan, Goyal, Torabi, Bashlykov, Bogoychev, Chatterji, Zhang, Duchenne, \c{C}elebi, Alrassy, Zhang, Li, Vasic, Weng, Bhargava, Dubal, Krishnan, Koura, Xu, He, Dong, Srinivasan, Ganapathy, Calderer, Cabral, Stojnic, Raileanu, Maheswari, Girdhar, Patel, Sauvestre, Polidoro, Sumbaly, Taylor, Silva, Hou, Wang, Hosseini, Chennabasappa, Singh, Bell, Kim, Edunov, Nie, Narang, Raparthy, Shen, Wan, Bhosale, Zhang, Vandenhende, Batra, Whitman, Sootla, Collot, Gururangan, Borodinsky, Herman, Fowler, Sheasha, Georgiou, Scialom, Speckbacher, Mihaylov, Xiao, Karn, Goswami, Gupta, Ramanathan, Kerkez, Gonguet, Do, Vogeti, Albiero, Petrovic, Chu, Xiong, Fu, Meers, Martinet, Wang, Wang, Tan, Xia, Xie, Jia, Wang, Goldschlag, Gaur, Babaei, Wen, Song, Zhang, Li, Mao, Coudert, Yan, Chen, Papakipos, Singh, Srivastava, Jain, Kelsey, Shajnfeld, Gangidi, Victoria, Goldstand, Menon, Sharma, Boesenberg,
  Baevski, Feinstein, Kallet, Sangani, Teo, Yunus, Lupu, Alvarado, Caples, Gu, Ho, Poulton, Ryan, Ramchandani, Dong, Franco, Goyal, Saraf, Chowdhury, Gabriel, Bharambe, Eisenman, Yazdan, James, Maurer, Leonhardi, Huang, Loyd, De~Paola, Paranjape, Liu, Wu, Ni, Hancock, Wasti, Spence, Stojkovic, Gamido, Montalvo, Parker, Burton, Mejia, Liu, Wang, Kim, Zhou, Hu, Chu, Cai, Tindal, Feichtenhofer, Gao, Civin, Beaty, Kreymer, Li, Adkins, Xu, Testuggine, David, Parikh, Liskovich, Foss, Wang, Le, Holland, Dowling, Jamil, Montgomery, Presani, Hahn, Wood, Le, Brinkman, Arcaute, Dunbar, Smothers, Sun, Kreuk, Tian, Kokkinos, Ozgenel, Caggioni, Kanayet, Seide, Florez, Schwarz, Badeer, Swee, Halpern, Herman, Sizov, {Guangyi}, {Zhang}, Lakshminarayanan, Inan, Shojanazeri, Zou, Wang, Zha, Habeeb, Rudolph, Suk, Aspegren, Goldman, Zhan, Damlaj, Molybog, Tufanov, Leontiadis, Veliche, Gat, Weissman, Geboski, Kohli, Lam, Asher, Gaya, Marcus, Tang, Chan, Zhen, Reizenstein, Teboul, Zhong, Jin, Yang, Cummings, Carvill, Shepard,
  McPhie, Torres, Ginsburg, Wang, Wu, U, Saxena, Khandelwal, Zand, Matosich, Veeraraghavan, Michelena, Li, Jagadeesh, Huang, Chawla, Huang, Chen, Garg, A, Silva, Bell, Zhang, Guo, Yu, Moshkovich, Wehrstedt, Khabsa, Avalani, Bhatt, Mankus, Hasson, Lennie, Reso, Groshev, Naumov, Lathi, Keneally, Liu, Seltzer, Valko, Restrepo, Patel, Vyatskov, Samvelyan, Clark, Macey, Wang, Hermoso, Metanat, Rastegari, Bansal, Santhanam, Parks, White, Bawa, Singhal, Egebo, Usunier, Mehta, Laptev, Dong, Cheng, Chernoguz, Hart, Salpekar, Kalinli, Kent, Parekh, Saab, Balaji, Rittner, Bontrager, Roux, Dollar, Zvyagina, Ratanchandani, Yuvraj, Liang, Alao, Rodriguez, Ayub, Murthy, Nayani, Mitra, Parthasarathy, Li, Hogan, Battey, Wang, Howes, Rinott, Mehta, Siby, Bondu, Datta, Chugh, Hunt, Dhillon, Sidorov, Pan, Mahajan, Verma, Yamamoto, Ramaswamy, Lindsay, Lindsay, Feng, Lin, Zha, Patil, Shankar, Zhang, Zhang, Wang, Agarwal, Sajuyigbe, Chintala, Max, Chen, Kehoe, Satterfield, Govindaprasad, Gupta, Deng, Cho, Virk, Subramanian,
  Choudhury, Goldman, Remez, Glaser, Best, Koehler, Robinson, Li, Zhang, Matthews, Chou, Shaked, Vontimitta, Ajayi, Montanez, Mohan, Kumar, Mangla, Ionescu, Poenaru, Mihailescu, Ivanov, Li, Wang, Jiang, Bouaziz, Constable, Tang, Wu, Wang, Wu, Gao, Kleinman, Chen, Hu, Jia, Qi, Li, Zhang, Zhang, Adi, Nam, {Yu}, {Wang}, Zhao, Hao, Qian, Li, He, Rait, DeVito, Rosnbrick, Wen, Yang, Zhao, and Ma]{Grattafiori2024-rd}
Aaron Grattafiori, Abhimanyu Dubey, Abhinav Jauhri, Abhinav Pandey, Abhishek Kadian, Ahmad Al-Dahle, Aiesha Letman, Akhil Mathur, Alan Schelten, Alex Vaughan, Amy Yang, Angela Fan, Anirudh Goyal, Anthony Hartshorn, Aobo Yang, Archi Mitra, Archie Sravankumar, Artem Korenev, Arthur Hinsvark, Arun Rao, Aston Zhang, Aurelien Rodriguez, Austen Gregerson, Ava Spataru, Baptiste Roziere, Bethany Biron, Binh Tang, Bobbie Chern, Charlotte Caucheteux, Chaya Nayak, Chloe Bi, Chris Marra, Chris McConnell, Christian Keller, Christophe Touret, Chunyang Wu, Corinne Wong, Cristian~Canton Ferrer, Cyrus Nikolaidis, Damien Allonsius, Daniel Song, Danielle Pintz, Danny Livshits, Danny Wyatt, David Esiobu, Dhruv Choudhary, Dhruv Mahajan, Diego Garcia-Olano, Diego Perino, Dieuwke Hupkes, Egor Lakomkin, Ehab AlBadawy, Elina Lobanova, Emily Dinan, Eric~Michael Smith, Filip Radenovic, Francisco Guzm\'{a}n, Frank Zhang, Gabriel Synnaeve, Gabrielle Lee, Georgia~Lewis Anderson, Govind Thattai, Graeme Nail, Gregoire Mialon, Guan Pang,
  Guillem Cucurell, Hailey Nguyen, Hannah Korevaar, Hu~Xu, Hugo Touvron, Iliyan Zarov, Imanol~Arrieta Ibarra, Isabel Kloumann, Ishan Misra, Ivan Evtimov, Jack Zhang, Jade Copet, Jaewon Lee, Jan Geffert, Jana Vranes, Jason Park, Jay Mahadeokar, Jeet Shah, Jelmer van~der Linde, Jennifer Billock, Jenny Hong, Jenya Lee, Jeremy Fu, Jianfeng Chi, Jianyu Huang, Jiawen Liu, Jie Wang, Jiecao Yu, Joanna Bitton, Joe Spisak, Jongsoo Park, Joseph Rocca, Joshua Johnstun, Joshua Saxe, Junteng Jia, Kalyan~Vasuden Alwala, Karthik Prasad, Kartikeya Upasani, Kate Plawiak, Ke~Li, Kenneth Heafield, Kevin Stone, Khalid El-Arini, Krithika Iyer, Kshitiz Malik, Kuenley Chiu, Kunal Bhalla, Kushal Lakhotia, Lauren Rantala-Yeary, Laurens van~der Maaten, Lawrence Chen, Liang Tan, Liz Jenkins, Louis Martin, Lovish Madaan, Lubo Malo, Lukas Blecher, Lukas Landzaat, Luke de~Oliveira, Madeline Muzzi, Mahesh Pasupuleti, Mannat Singh, Manohar Paluri, Marcin Kardas, Maria Tsimpoukelli, Mathew Oldham, Mathieu Rita, Maya Pavlova, Melanie Kambadur,
  Mike Lewis, Min Si, Mitesh~Kumar Singh, Mona Hassan, Naman Goyal, Narjes Torabi, Nikolay Bashlykov, Nikolay Bogoychev, Niladri Chatterji, Ning Zhang, Olivier Duchenne, Onur \c{C}elebi, Patrick Alrassy, Pengchuan Zhang, Pengwei Li, Petar Vasic, Peter Weng, Prajjwal Bhargava, Pratik Dubal, Praveen Krishnan, Punit~Singh Koura, Puxin Xu, Qing He, Qingxiao Dong, Ragavan Srinivasan, Raj Ganapathy, Ramon Calderer, Ricardo~Silveira Cabral, Robert Stojnic, Roberta Raileanu, Rohan Maheswari, Rohit Girdhar, Rohit Patel, Romain Sauvestre, Ronnie Polidoro, Roshan Sumbaly, Ross Taylor, Ruan Silva, Rui Hou, Rui Wang, Saghar Hosseini, Sahana Chennabasappa, Sanjay Singh, Sean Bell, Seohyun~Sonia Kim, Sergey Edunov, Shaoliang Nie, Sharan Narang, Sharath Raparthy, Sheng Shen, Shengye Wan, Shruti Bhosale, Shun Zhang, Simon Vandenhende, Soumya Batra, Spencer Whitman, Sten Sootla, Stephane Collot, Suchin Gururangan, Sydney Borodinsky, Tamar Herman, Tara Fowler, Tarek Sheasha, Thomas Georgiou, Thomas Scialom, Tobias Speckbacher,
  Todor Mihaylov, Tong Xiao, Ujjwal Karn, Vedanuj Goswami, Vibhor Gupta, Vignesh Ramanathan, Viktor Kerkez, Vincent Gonguet, Virginie Do, Vish Vogeti, V\'{\i}tor Albiero, Vladan Petrovic, Weiwei Chu, Wenhan Xiong, Wenyin Fu, Whitney Meers, Xavier Martinet, Xiaodong Wang, Xiaofang Wang, Xiaoqing~Ellen Tan, Xide Xia, Xinfeng Xie, Xuchao Jia, Xuewei Wang, Yaelle Goldschlag, Yashesh Gaur, Yasmine Babaei, Yi~Wen, Yiwen Song, Yuchen Zhang, Yue Li, Yuning Mao, Zacharie~Delpierre Coudert, Zheng Yan, Zhengxing Chen, Zoe Papakipos, Aaditya Singh, Aayushi Srivastava, Abha Jain, Adam Kelsey, Adam Shajnfeld, Adithya Gangidi, Adolfo Victoria, Ahuva Goldstand, Ajay Menon, Ajay Sharma, Alex Boesenberg, Alexei Baevski, Allie Feinstein, Amanda Kallet, Amit Sangani, Amos Teo, Anam Yunus, Andrei Lupu, Andres Alvarado, Andrew Caples, Andrew Gu, Andrew Ho, Andrew Poulton, Andrew Ryan, Ankit Ramchandani, Annie Dong, Annie Franco, Anuj Goyal, Aparajita Saraf, Arkabandhu Chowdhury, Ashley Gabriel, Ashwin Bharambe, Assaf Eisenman,
  Azadeh Yazdan, Beau James, Ben Maurer, Benjamin Leonhardi, Bernie Huang, Beth Loyd, Beto De~Paola, Bhargavi Paranjape, Bing Liu, Bo~Wu, Boyu Ni, Braden Hancock, Bram Wasti, Brandon Spence, Brani Stojkovic, Brian Gamido, Britt Montalvo, Carl Parker, Carly Burton, Catalina Mejia, Ce~Liu, Changhan Wang, Changkyu Kim, Chao Zhou, Chester Hu, Ching-Hsiang Chu, Chris Cai, Chris Tindal, Christoph Feichtenhofer, Cynthia Gao, Damon Civin, Dana Beaty, Daniel Kreymer, Daniel Li, David Adkins, David Xu, Davide Testuggine, Delia David, Devi Parikh, Diana Liskovich, Didem Foss, Dingkang Wang, Duc Le, Dustin Holland, Edward Dowling, Eissa Jamil, Elaine Montgomery, Eleonora Presani, Emily Hahn, Emily Wood, Eric-Tuan Le, Erik Brinkman, Esteban Arcaute, Evan Dunbar, Evan Smothers, Fei Sun, Felix Kreuk, Feng Tian, Filippos Kokkinos, Firat Ozgenel, Francesco Caggioni, Frank Kanayet, Frank Seide, Gabriela~Medina Florez, Gabriella Schwarz, Gada Badeer, Georgia Swee, Gil Halpern, Grant Herman, Grigory Sizov, {Guangyi}, {Zhang},
  Guna Lakshminarayanan, Hakan Inan, Hamid Shojanazeri, Han Zou, Hannah Wang, Hanwen Zha, Haroun Habeeb, Harrison Rudolph, Helen Suk, Henry Aspegren, Hunter Goldman, Hongyuan Zhan, Ibrahim Damlaj, Igor Molybog, Igor Tufanov, Ilias Leontiadis, Irina-Elena Veliche, Itai Gat, Jake Weissman, James Geboski, James Kohli, Janice Lam, Japhet Asher, Jean-Baptiste Gaya, Jeff Marcus, Jeff Tang, Jennifer Chan, Jenny Zhen, Jeremy Reizenstein, Jeremy Teboul, Jessica Zhong, Jian Jin, Jingyi Yang, Joe Cummings, Jon Carvill, Jon Shepard, Jonathan McPhie, Jonathan Torres, Josh Ginsburg, Junjie Wang, Kai Wu, Kam~Hou U, Karan Saxena, Kartikay Khandelwal, Katayoun Zand, Kathy Matosich, Kaushik Veeraraghavan, Kelly Michelena, Keqian Li, Kiran Jagadeesh, Kun Huang, Kunal Chawla, Kyle Huang, Lailin Chen, Lakshya Garg, Lavender A, Leandro Silva, Lee Bell, Lei Zhang, Liangpeng Guo, Licheng Yu, Liron Moshkovich, Luca Wehrstedt, Madian Khabsa, Manav Avalani, Manish Bhatt, Martynas Mankus, Matan Hasson, Matthew Lennie, Matthias Reso,
  Maxim Groshev, Maxim Naumov, Maya Lathi, Meghan Keneally, Miao Liu, Michael~L Seltzer, Michal Valko, Michelle Restrepo, Mihir Patel, Mik Vyatskov, Mikayel Samvelyan, Mike Clark, Mike Macey, Mike Wang, Miquel~Jubert Hermoso, Mo~Metanat, Mohammad Rastegari, Munish Bansal, Nandhini Santhanam, Natascha Parks, Natasha White, Navyata Bawa, Nayan Singhal, Nick Egebo, Nicolas Usunier, Nikhil Mehta, Nikolay~Pavlovich Laptev, Ning Dong, Norman Cheng, Oleg Chernoguz, Olivia Hart, Omkar Salpekar, Ozlem Kalinli, Parkin Kent, Parth Parekh, Paul Saab, Pavan Balaji, Pedro Rittner, Philip Bontrager, Pierre Roux, Piotr Dollar, Polina Zvyagina, Prashant Ratanchandani, Pritish Yuvraj, Qian Liang, Rachad Alao, Rachel Rodriguez, Rafi Ayub, Raghotham Murthy, Raghu Nayani, Rahul Mitra, Rangaprabhu Parthasarathy, Raymond Li, Rebekkah Hogan, Robin Battey, Rocky Wang, Russ Howes, Ruty Rinott, Sachin Mehta, Sachin Siby, Sai~Jayesh Bondu, Samyak Datta, Sara Chugh, Sara Hunt, Sargun Dhillon, Sasha Sidorov, Satadru Pan, Saurabh Mahajan,
  Saurabh Verma, Seiji Yamamoto, Sharadh Ramaswamy, Shaun Lindsay, Shaun Lindsay, Sheng Feng, Shenghao Lin, Shengxin~Cindy Zha, Shishir Patil, Shiva Shankar, Shuqiang Zhang, Shuqiang Zhang, Sinong Wang, Sneha Agarwal, Soji Sajuyigbe, Soumith Chintala, Stephanie Max, Stephen Chen, Steve Kehoe, Steve Satterfield, Sudarshan Govindaprasad, Sumit Gupta, Summer Deng, Sungmin Cho, Sunny Virk, Suraj Subramanian, Sy~Choudhury, Sydney Goldman, Tal Remez, Tamar Glaser, Tamara Best, Thilo Koehler, Thomas Robinson, Tianhe Li, Tianjun Zhang, Tim Matthews, Timothy Chou, Tzook Shaked, Varun Vontimitta, Victoria Ajayi, Victoria Montanez, Vijai Mohan, Vinay~Satish Kumar, Vishal Mangla, Vlad Ionescu, Vlad Poenaru, Vlad~Tiberiu Mihailescu, Vladimir Ivanov, Wei Li, Wenchen Wang, Wenwen Jiang, Wes Bouaziz, Will Constable, Xiaocheng Tang, Xiaojian Wu, Xiaolan Wang, Xilun Wu, Xinbo Gao, Yaniv Kleinman, Yanjun Chen, Ye~Hu, Ye~Jia, Ye~Qi, Yenda Li, Yilin Zhang, Ying Zhang, Yossi Adi, Youngjin Nam, {Yu}, {Wang}, Yu~Zhao, Yuchen Hao,
  Yundi Qian, Yunlu Li, Yuzi He, Zach Rait, Zachary DeVito, Zef Rosnbrick, Zhaoduo Wen, Zhenyu Yang, Zhiwei Zhao, and Zhiyu Ma.
\newblock The llama 3 herd of models.
\newblock \emph{arXiv [cs.AI]}, 31~July 2024.

\bibitem[{Qwen} et~al.(2024){Qwen}, Yang, Yang, Zhang, Hui, Zheng, Yu, Li, Liu, Huang, Wei, Lin, Yang, Tu, Zhang, Yang, Yang, Zhou, Lin, Dang, Lu, Bao, Yang, Yu, Li, Xue, Zhang, Zhu, Men, Lin, Li, Tang, Xia, Ren, Ren, Fan, Su, Zhang, Wan, Liu, Cui, Zhang, and Qiu]{Qwen2024-mg}
{Qwen}, An~Yang, Baosong Yang, Beichen Zhang, Binyuan Hui, Bo~Zheng, Bowen Yu, Chengyuan Li, Dayiheng Liu, Fei Huang, Haoran Wei, Huan Lin, Jian Yang, Jianhong Tu, Jianwei Zhang, Jianxin Yang, Jiaxi Yang, Jingren Zhou, Junyang Lin, Kai Dang, Keming Lu, Keqin Bao, Kexin Yang, Le~Yu, Mei Li, Mingfeng Xue, Pei Zhang, Qin Zhu, Rui Men, Runji Lin, Tianhao Li, Tianyi Tang, Tingyu Xia, Xingzhang Ren, Xuancheng Ren, Yang Fan, Yang Su, Yichang Zhang, Yu~Wan, Yuqiong Liu, Zeyu Cui, Zhenru Zhang, and Zihan Qiu.
\newblock {Qwen2}.5 technical report.
\newblock \emph{arXiv [cs.CL]}, 19~December 2024.

\bibitem[Abdin et~al.(2024)Abdin, Aneja, Awadalla, Awadallah, Awan, Bach, Bahree, Bakhtiari, Bao, Behl, Benhaim, Bilenko, Bjorck, Bubeck, Cai, Cai, Chaudhary, Chen, Chen, Chen, Chen, Chen, Cheng, Chopra, Dai, Dixon, Eldan, Fragoso, Gao, Gao, Gao, Garg, Del~Giorno, Goswami, Gunasekar, Haider, Hao, Hewett, Hu, Huynh, Iter, Jacobs, Javaheripi, Jin, Karampatziakis, Kauffmann, Khademi, Kim, Kim, Kurilenko, Lee, Lee, Li, Li, Liang, Liden, Lin, Lin, Liu, Liu, Liu, Liu, Liu, Luo, Madan, Mahmoudzadeh, Majercak, Mazzola, Mendes, Mitra, Modi, Nguyen, Norick, Patra, Perez-Becker, Portet, Pryzant, Qin, Radmilac, Ren, de~Rosa, Rosset, Roy, Ruwase, Saarikivi, Saied, Salim, Santacroce, Shah, Shang, Sharma, Shen, Shukla, Song, Tanaka, Tupini, Vaddamanu, Wang, Wang, Wang, Wang, Wang, Wang, Ward, Wen, Witte, Wu, Wu, Wyatt, Xiao, Xu, Xu, Xu, Xue, Yadav, Yang, Yang, Yang, Yang, Yu, Yuan, Zhang, Zhang, Zhang, Zhang, Zhang, Zhang, Zhang, and Zhou]{Abdin2024-uo}
Marah Abdin, Jyoti Aneja, Hany Awadalla, Ahmed Awadallah, Ammar~Ahmad Awan, Nguyen Bach, Amit Bahree, Arash Bakhtiari, Jianmin Bao, Harkirat Behl, Alon Benhaim, Misha Bilenko, Johan Bjorck, S\'{e}bastien Bubeck, Martin Cai, Qin Cai, Vishrav Chaudhary, Dong Chen, Dongdong Chen, Weizhu Chen, Yen-Chun Chen, Yi-Ling Chen, Hao Cheng, Parul Chopra, Xiyang Dai, Matthew Dixon, Ronen Eldan, Victor Fragoso, Jianfeng Gao, Mei Gao, Min Gao, Amit Garg, Allie Del~Giorno, Abhishek Goswami, Suriya Gunasekar, Emman Haider, Junheng Hao, Russell~J Hewett, Wenxiang Hu, Jamie Huynh, Dan Iter, Sam~Ade Jacobs, Mojan Javaheripi, Xin Jin, Nikos Karampatziakis, Piero Kauffmann, Mahoud Khademi, Dongwoo Kim, Young~Jin Kim, Lev Kurilenko, James~R Lee, Yin~Tat Lee, Yuanzhi Li, Yunsheng Li, Chen Liang, Lars Liden, Xihui Lin, Zeqi Lin, Ce~Liu, Liyuan Liu, Mengchen Liu, Weishung Liu, Xiaodong Liu, Chong Luo, Piyush Madan, Ali Mahmoudzadeh, David Majercak, Matt Mazzola, Caio C\'{e}sar~Teodoro Mendes, Arindam Mitra, Hardik Modi, Anh Nguyen,
  Brandon Norick, Barun Patra, Daniel Perez-Becker, Thomas Portet, Reid Pryzant, Heyang Qin, Marko Radmilac, Liliang Ren, Gustavo de~Rosa, Corby Rosset, Sambudha Roy, Olatunji Ruwase, Olli Saarikivi, Amin Saied, Adil Salim, Michael Santacroce, Shital Shah, Ning Shang, Hiteshi Sharma, Yelong Shen, Swadheen Shukla, Xia Song, Masahiro Tanaka, Andrea Tupini, Praneetha Vaddamanu, Chunyu Wang, Guanhua Wang, Lijuan Wang, Shuohang Wang, Xin Wang, Yu~Wang, Rachel Ward, Wen Wen, Philipp Witte, Haiping Wu, Xiaoxia Wu, Michael Wyatt, Bin Xiao, Can Xu, Jiahang Xu, Weijian Xu, Jilong Xue, Sonali Yadav, Fan Yang, Jianwei Yang, Yifan Yang, Ziyi Yang, Donghan Yu, Lu~Yuan, Chenruidong Zhang, Cyril Zhang, Jianwen Zhang, Li~Lyna Zhang, Yi~Zhang, Yue Zhang, Yunan Zhang, and Xiren Zhou.
\newblock Phi-3 technical report: A highly capable language model locally on your phone.
\newblock \emph{arXiv [cs.CL]}, 22~April 2024.

\bibitem[Han et~al.(2023)Han, Adams, Papaioannou, Grundmann, Oberhauser, L{\"o}ser, Truhn, and Bressem]{han2023medalpaca}
Tianyu Han, Lisa~C Adams, Jens-Michalis Papaioannou, Paul Grundmann, Tom Oberhauser, Alexander L{\"o}ser, Daniel Truhn, and Keno~K Bressem.
\newblock Medalpaca--an open-source collection of medical conversational ai models and training data.
\newblock \emph{arXiv preprint arXiv:2304.08247}, 2023.

\bibitem[Alsentzer et~al.(2019)Alsentzer, Murphy, Boag, Weng, Jindi, Naumann, and McDermott]{alsentzer2019publicly}
Emily Alsentzer, John Murphy, William Boag, Wei-Hung Weng, Di~Jindi, Tristan Naumann, and Matthew McDermott.
\newblock Publicly available clinical bert embeddings.
\newblock In \emph{Proceedings of the 2nd clinical natural language processing workshop}, pages 72--78, 2019.

\bibitem[Team et~al.(2024)Team, Anil, Borgeaud, Wu, Alayrac, Yu, Soricut, Schalkwyk, Dai, Hauth, et~al.]{team2024gemini}
Gemini Team, R~Anil, S~Borgeaud, Y~Wu, JB~Alayrac, J~Yu, R~Soricut, J~Schalkwyk, AM~Dai, A~Hauth, et~al.
\newblock Gemini: A family of highly capable multimodal models, 2024.
\newblock \emph{arXiv preprint arXiv:2312.11805}, 10, 2024.

\bibitem[Fan et~al.(2025)Fan, Liang, Wu, Zhang, Wang, and Xie]{Fan2025-nc}
Ziqing Fan, Cheng Liang, Chaoyi Wu, Ya~Zhang, Yanfeng Wang, and Weidi Xie.
\newblock {ChestX}-reasoner: Advancing radiology foundation models with reasoning through step-by-step verification.
\newblock \emph{arXiv [cs.AI]}, 29~April 2025.

\bibitem[Gu et~al.(2024)Gu, Jiang, Shi, Tan, Zhai, Xu, Li, Shen, Ma, Liu, Wang, Zhang, Wang, Gao, Ni, and Guo]{gu2024survey}
Jiawei Gu, Xuhui Jiang, Zhichao Shi, Hexiang Tan, Xuehao Zhai, Chengjin Xu, Wei Li, Yinghan Shen, Shengjie Ma, Honghao Liu, Saizhuo Wang, Kun Zhang, Yuanzhuo Wang, Wen Gao, Lionel Ni, and Jian Guo.
\newblock A survey on {LLM}-as-a-judge.
\newblock \emph{arXiv [cs.CL]}, 23~November 2024.

\bibitem[Kirkpatrick et~al.(2017)Kirkpatrick, Pascanu, Rabinowitz, Veness, Desjardins, Rusu, Milan, Quan, Ramalho, Grabska-Barwinska, Hassabis, Clopath, Kumaran, and Hadsell]{kirkpatrick2017overcoming}
James Kirkpatrick, Razvan Pascanu, Neil Rabinowitz, Joel Veness, Guillaume Desjardins, Andrei~A Rusu, Kieran Milan, John Quan, Tiago Ramalho, Agnieszka Grabska-Barwinska, Demis Hassabis, Claudia Clopath, Dharshan Kumaran, and Raia Hadsell.
\newblock Overcoming catastrophic forgetting in neural networks.
\newblock \emph{Proc. Natl. Acad. Sci. U. S. A.}, 114\penalty0 (13):\penalty0 3521--3526, 28~March 2017.
\newblock ISSN 0027-8424,1091-6490.
\newblock \doi{10.1073/pnas.1611835114}.

\bibitem[Tadros et~al.(2022)Tadros, Krishnan, Ramyaa, and Bazhenov]{tadros2022sleep-like}
Timothy Tadros, Giri~P Krishnan, Ramyaa Ramyaa, and Maxim Bazhenov.
\newblock Sleep-like unsupervised replay reduces catastrophic forgetting in artificial neural networks.
\newblock \emph{Nat. Commun.}, 13\penalty0 (1):\penalty0 7742, 15~December 2022.
\newblock ISSN 2041-1723.
\newblock \doi{10.1038/s41467-022-34938-7}.

\bibitem[Johnson et~al.(2019)Johnson, Pollard, Greenbaum, Lungren, Deng, Peng, Lu, Mark, Berkowitz, and Horng]{Johnson2019-eb}
Alistair E~W Johnson, Tom~J Pollard, Nathaniel~R Greenbaum, Matthew~P Lungren, Chih-Ying Deng, Yifan Peng, Zhiyong Lu, Roger~G Mark, Seth~J Berkowitz, and Steven Horng.
\newblock {MIMIC}-{CXR}-{JPG}, a large publicly available database of labeled chest radiographs.
\newblock \emph{arXiv [cs.CV]}, 21~January 2019.

\bibitem[Waisberg et~al.(2023)Waisberg, Ong, Masalkhi, Kamran, Zaman, Sarker, Lee, and Tavakkoli]{Waisberg2023-fu}
Ethan Waisberg, Joshua Ong, Mouayad Masalkhi, Sharif~Amit Kamran, Nasif Zaman, Prithul Sarker, Andrew~G Lee, and Alireza Tavakkoli.
\newblock {GPT}-4: a new era of artificial intelligence in medicine.
\newblock \emph{Ir. J. Med. Sci.}, 192\penalty0 (6):\penalty0 3197--3200, December 2023.
\newblock ISSN 0021-1265,1863-4362.
\newblock \doi{10.1007/s11845-023-03377-8}.

\bibitem[Hu et~al.(2021)Hu, Shen, Wallis, Allen-Zhu, Li, Wang, and Chen]{Hu2021-qq}
J~E Hu, Yelong Shen, Phillip Wallis, Zeyuan Allen-Zhu, Yuanzhi Li, Shean Wang, and Weizhu Chen.
\newblock {LoRA}: Low-rank adaptation of large language models.
\newblock \emph{Int Conf Learn Represent}, abs/2106.09685, 17~June 2021.

\bibitem[von Werra et~al.(2020)von Werra, Belkada, Tunstall, Beeching, Thrush, Lambert, Huang, Rasul, and Gallouédec]{von-Werra2020-ca}
Leandro von Werra, Younes Belkada, Lewis Tunstall, Edward Beeching, Tristan Thrush, Nathan Lambert, Shengyi Huang, Kashif Rasul, and Quentin Gallouédec.
\newblock Trl: Transformer reinforcement learning.
\newblock \url{https://github.com/huggingface/trl}, 2020.

\bibitem[Dao et~al.(2022)Dao, Fu, Ermon, Rudra, and R'e]{Dao2022-wv}
Tri Dao, Daniel~Y Fu, Stefano Ermon, A~Rudra, and Christopher R'e.
\newblock {FlashAttention}: Fast and memory-efficient exact attention with {IO}-awareness.
\newblock \emph{Neural Inf Process Syst}, abs/2205.14135:\penalty0 16344--16359, 27~May 2022.

\end{thebibliography}

\newpage
\appendix
\setcounter{table}{0}
\setcounter{figure}{0}
\setcounter{prompt}{0}
\renewcommand\figurename{Supplementary Figure} 
\renewcommand\tablename{Supplementary Table}
\renewcommand{\boxedname}{Supplementary Prompt}
\renewcommand{\thefigure}{S\arabic{figure}}
\renewcommand{\thetable}{S\arabic{table}}
\renewcommand{\theprompt}{S\arabic{prompt}}

\section*{Supplementary materials}
\singlespacing



%



\begin{table}[!phtb]
\caption{Prediction accuracy of individual predictions for MIMIC-CXR dataset}
\label{stab:individual_mimic}
\centering
\footnotesize

\begin{tabular}{l lc c c c cc c c c c}
\toprule
 &  & \multicolumn{5}{c}{Precision} & \multicolumn{5}{c}{Recall} \\
\cmidrule(rl){3-7}\cmidrule(rl){8-12}
Model & Method &  1 & 2 & 3 & 4 & 5 & 1 & 2 & 3 & 4 & 5 \\
\midrule
Llama & Baseline   & 0.045 & 0.327 & 0.318 & 0.215 & 0.353 & 0.041 & 0.171 & 0.394 & 0.218 & 0.218 \\
      & SFT        & 0.828 & 0.796 & 0.808 & 0.829 & 0.801 & 0.699 & 0.689 & 0.699 & 0.705 & 0.710 \\
      & SFT+GRPO   & 0.721 & 0.723 & 0.766 & 0.779 & 0.785 & 0.736 & 0.731 & 0.731 & 0.694 & 0.736 \\
\midrule
Qwen  & Baseline   & 0.347 & 0.452 & 0.390 & 0.371 & 0.178 & 0.218 & 0.098 & 0.202 & 0.187 & 0.124 \\
      & SFT        & 0.787 & 0.819 & 0.803 & 0.806 & 0.828 & 0.554 & 0.539 & 0.570 & 0.539 & 0.523 \\
      & SFT+GRPO   & 0.800 & 0.791 & 0.828 & 0.776 & 0.799 & 0.601 & 0.648 & 0.648 & 0.627 & 0.596 \\
\midrule
Phi4  & Baseline   & 0.333 & 0.353 & 0.400 & 0.372 & 0.453 & 0.244 & 0.378 & 0.383 & 0.378 & 0.399 \\
      & SFT        & 0.695 & 0.674 & 0.669 & 0.667 & 0.642 & 0.591 & 0.482 & 0.482 & 0.477 & 0.456 \\
      & SFT+GRPO   & 0.730 & 0.784 & 0.711 & 0.736 & 0.781 & 0.658 & 0.658 & 0.637 & 0.663 & 0.627 \\
\midrule
 & & \multicolumn{5}{c}{F1} & \multicolumn{3}{c}{Average} \\
 \cmidrule(rl){3-7}\cmidrule(rl){8-10}
 &  & 1 & 2 & 3 & 4 & 5 & Precision & Recall & F1 \\
\cmidrule(rl){1-10}
Llama & Baseline   & 0.043 & 0.225 & 0.352 & 0.216 & 0.270 & 0.251 & 0.208 & 0.221 \\
      & SFT        & 0.758 & 0.739 & 0.750 & 0.762 & 0.753 & 0.812 & 0.700 & 0.752 \\
      & SFT+GRPO   & 0.728 & 0.727 & 0.748 & 0.734 & 0.760 & 0.754 & 0.725 & 0.740 \\
\cmidrule(rl){1-10}
Qwen  & Baseline   & 0.268 & 0.161 & 0.266 & 0.249 & 0.146 & 0.347 & 0.165 & 0.218 \\
      & SFT        & 0.650 & 0.650 & 0.667 & 0.646 & 0.641 & 0.808 & 0.545 & 0.651 \\
      & SFT+GRPO   & 0.686 & 0.712 & 0.727 & 0.694 & 0.683 & 0.798 & 0.624 & 0.700 \\
\cmidrule(rl){1-10}
Phi4  & Baseline   & 0.282 & 0.365 & 0.391 & 0.375 & 0.424 & 0.382 & 0.356 & 0.367 \\
      & SFT        & 0.639 & 0.562 & 0.560 & 0.556 & 0.533 & 0.669 & 0.497 & 0.570 \\
      & SFT+GRPO   & 0.692 & 0.715 & 0.672 & 0.698 & 0.696 & 0.748 & 0.648 & 0.695 \\
\bottomrule
\end{tabular}

\end{table}

\newpage

\begin{table}[!th]
\caption{Results of prediction accuracy on all variants.}
\label{stab:variants}
\centering
\tiny
\begin{tabular}{llccc}
\toprule
Dataset & Model & Baseline & SFT & SFT + GRPO \\
\midrule
\rowcolor{graybg} \multicolumn{5}{c}{Micro precision} \\ 
MIMIC & Qwen-3B & 0.452 [0.311, 0.586] & 0.811 [0.744, 0.871] & 0.824 [0.764, 0.881] \\ 
~ & Llama & 0.684 [0.528, 0.824] & 0.873 [0.825, 0.919] & 0.882 [0.831, 0.931] \\ 
~ & Phi3 & 0.511 [0.431, 0.588] & 0.845 [0.776, 0.908] & 0.791 [0.731, 0.849] \\ 
& Qwen-7B & 0.651 [0.557, 0.740] & 0.691 [0.636, 0.744] & 0.769 [0.712, 0.824] \\ 
& Medalpaca & 0.046 [0.000, 0.111] & 0.770 [0.723, 0.816] & 0.881 [0.839, 0.924] \\ 
~ & GPT4-o & 0.758 [0.707, 0.813] & - & - \\ 
~ & BioClinicalBERT & 0.870 [0.815, 0.929] & - & - \\ 
\midrule
NIH-CXR & Qwen-3B & 0.362 [0.255, 0.462] & 0.907 [0.846, 0.960] & 0.875 [0.804, 0.940] \\ 
~ & Llama & 0.286 [0.160, 0.412] & 0.929 [0.856, 0.988] & 0.870 [0.805, 0.929] \\ 
~ & Phi3 & 0.429 [0.341, 0.519] & 0.795 [0.705, 0.885] & 0.829 [0.747, 0.902] \\ 
&Qwen-7B & 0.602 [0.480, 0.714] & 0.827 [0.768, 0.883] & 0.868 [0.811, 0.922] \\ 
& Medalpaca & 0.122 [0.018, 0.222] & 0.929 [0.868, 0.979] &  0.940[0.892, 0.981]\\ 
~ & GPT4-o & 0.776 [0.713, 0.833] & - & - \\ 
~ & BioClinicalBERT & 0.972 [0.928, 1.000] & - & - \\ 
\midrule
MIDRC & Qwen-3B & 0.298 [0.164, 0.427] & 0.827 [0.738, 0.908] & 0.767 [0.674, 0.845] \\ 
~ & Llama & 0.254 [0.167, 0.329] & 0.980 [0.949, 1.000] & 0.951 [0.911, 0.991] \\ 
~ & Phi3 & 0.243 [0.149, 0.329] & 0.582 [0.439, 0.646] & 0.560 [0.494, 0.714] \\ 
& Qwen-7B & 0.327 [0.242, 0.409] & 0.919 [0.873, 0.961] &  0.981 [0.955, 1.000]\\ 
& Medalpaca & 0.000 [0.000, 0.000] & 0.922 [0.875, 0.965] & 0.972 [0.941, 1.000] \\ 
~ & GPT4-o & 0.667 [0.636, 0.696] & - & - \\ 
~ & BioClinicalBERT &0.988 [0.963, 1.000] & - & - \\ 
\rowcolor{graybg} \multicolumn{5}{c}{Micro recall}\\ 
MIMIC & Qwen-3B & 0.145 [0.091, 0.208] & 0.534 [0.449, 0.610] & 0.653 [0.586, 0.722] \\ 
~ & Llama & 0.135 [0.088, 0.185] & 0.710 [0.646, 0.771] & 0.777 [0.718, 0.839] \\ 
~ & Phi3 & 0.368 [0.304, 0.436] & 0.508 [0.429, 0.582] & 0.648 [0.571, 0.720] \\ 
& Qwen-7B & 0.404 [0.335, 0.482] & 0.820 [0.769, 0.865]  & 0.809 [0.760, 0.856] \\ 
& Medalpaca & 0.026 [0.000, 0.067] & 0.830 [0.781, 0.881] & 0.813 [0.767, 0.861] \\ 
~ & GPT4-o & 0.907 [0.867, 0.944] & - & ~ \\ 
~ & BioClinicalBERT &0.628 [0.570, 0.694] & - & - \\ 
\midrule
NIH-CXR & Qwen-3B & 0.179 [0.121, 0.250] & 0.752 [0.677, 0.823] & 0.778 [0.701, 0.850] \\ 
~ & Llama & 0.137 [0.075, 0.218] & 0.778 [0.717, 0.848] & 0.855 [0.789, 0.917] \\ 
~ & Phi3 & 0.359 [0.282, 0.432] & 0.496 [0.407, 0.586] & 0.590 [0.500, 0.661] \\ 
& Qwen-7B & 0.342 [0.254, 0.424] &0.835 [0.771, 0.898]  & 0.817 [0.752, 0.882] \\ 
& Medalpaca & 0.081 [0.009, 0.170] & 0.742 [0.685, 0.802] &  0.758 [0.702, 0.812]\\ 
~ & GPT4-o & 0.949 [0.911, 0.983] & - & ~ \\ 
~ & BioClinicalBERT &0.575 [0.500, 0.650] & - & - \\ 
\midrule
MIDRC & Qwen-3B & 0.143 [0.071, 0.218] & 0.563 [0.480, 0.639] & 0.580 [0.587, 0.675] \\ 
~ & Llama & 0.143 [0.086, 0.208] & 0.824 [0.772, 0.878] & 0.983 [0.957, 1.000] \\ 
~ & Phi3 & 0.151 [0.090, 0.218] & 0.487 [0.275, 0.488] & 0.513 [0.328, 0.541] \\ 
& Qwen-7B & 0.210 [0.144, 0.275] & 0.958 [0.919, 0.991]  &  0.893 [0.843, 0.946]\\ 
& Medalpaca & 0.000 [0.000, 0.000] & 0.901 [0.847, 0.948] & 0.876 [0.827, 0.924] \\ 
~ & GPT4-o & 0.984 [0.958, 1.000] & - & - \\ 
~ & BioClinicalBERT &0.706 [0.655, 0.763]& - & - \\ 
\rowcolor{graybg} \multicolumn{5}{c}{Micro F1}\\ 
MIMIC & Qwen-3B & 0.220 [0.139, 0.296] & 0.644 [0.574, 0.710] & 0.729 [0.665, 0.786] \\ 
~ & Llama & 0.225 [0.157, 0.299] & 0.783 [0.732, 0.833] & 0.826 [0.775, 0.871] \\ 
~ & Phi3 & 0.428 [0.359, 0.494] & 0.635 [0.566, 0.702] & 0.712 [0.651, 0.773] \\ 
& Qwen-7B & 0.498 [0.420, 0.578] & 0.749 [0.706, 0.790] & 0.788 [0.748, 0.825] \\ 
& Medalpaca & 0.033 [0.000, 0.084] & 0.799 [0.758, 0.837] & 0.846 [0.811, 0.882] \\ 
~ & GPT4-o & 0.826 [0.787, 0.865] & - & - \\ 
~ & BioClinicalBERT &0.730 [0.683, 0.783]& - & - \\ 
\midrule
NIH-CXR & Qwen-3B & 0.240 [0.167, 0.320] & 0.822 [0.763, 0.872] & 0.824 [0.752, 0.880] \\ 
~ & Llama & 0.185 [0.098, 0.276] & 0.847 [0.796, 0.900] & 0.862 [0.813, 0.903] \\ 
~ & Phi3 & 0.391 [0.307, 0.467] & 0.611 [0.523, 0.694] & 0.689 [0.617, 0.745] \\ 
& Qwen-7B & 0.435 [0.337, 0.527] &0.830 [0.785, 0.872]  & 0.841 [0.800, 0.885] \\ 
& Medalpaca & 0.097 [0.012, 0.188] & 0.824 [0.781, 0.870] & 0.839 [0.798, 0.878] \\ 
~ & GPT4-o & 0.854 [0.811, 0.891] & - & - \\ 
~ & BioClinicalBERT &0.723 [0.659, 0.779]& - & - \\ 
\midrule
MIDRC & Qwen-3B & 0.193 [0.096, 0.295] & 0.670 [0.586, 0.745] & 0.661 [0.562, 0.748] \\ 
~ & Llama & 0.183 [0.112, 0.254] & 0.895 [0.863, 0.926] & 0.967 [0.942, 0.987] \\ 
~ & Phi3 & 0.186 [0.111, 0.263] & 0.509 [0.346, 0.545] & 0.535 [0.396, 0.610] \\
& Qwen-7B & 0.255 [0.179, 0.328] & 0.938 [0.910, 0.964] &  0.935 [0.905, 0.964]\\
& Medalpaca & 0.000 [0.000, 0.000] & 0.911 [0.875, 0.945] &  0.921 [0.886, 0.954] \\ 
~ & GPT4-o & 0.795 [0.770, 0.817] & - & - \\ 
~ & BioClinicalBERT &0.824 [0.784, 0.863]& - & - \\ 
\bottomrule
\end{tabular}
\end{table}

\newpage

\begin{table}[!phtb]
\caption{Reasoning performance comparison on baseline, SFT, SFT + GRPO.}
\label{stab:reasoning}
\centering
\begin{tabular}{llccc}
\toprule
Dataset & Model & Baseline & SFT & SFT + GRPO \\
\midrule
\rowcolor{graybg} \multicolumn{5}{c}{Reasoning recall} \\ 
MIMIC & Qwen-3B & 0.435 [0.357, 0.513] & 0.456 [0.376, 0.541] & 0.725 [0.660, 0.793] \\ 
~ & Llama & 0.171 [0.099, 0.246] & 0.000 [0.000, 0.000] & 0.647 [0.556, 0.739] \\ 
~ & Phi3 & 0.400 [0.330, 0.466] & 0.715 [0.660, 0.773] & 0.714 [0.644, 0.785] \\ 
\midrule
NIH-CXR & Qwen-3B & 0.462 [0.362, 0.573] & 0.650 [0.518, 0.779] & 0.792 [0.713, 0.870] \\ 
~ & Llama & 0.208 [0.131, 0.288] & 0.000 [0.000, 0.000] & 0.567 [0.430, 0.702] \\ 
~ & Phi3 & 0.525 [0.446, 0.603] & 0.625 [0.508, 0.735] & 0.692 [0.606, 0.773] \\ 
\midrule
MIDRC & Qwen-3B & 0.362 [0.290, 0.437] & 0.529 [0.419, 0.636] & 0.664 [0.559, 0.750] \\ 
~ & Llama & 0.277 [0.171, 0.378] & 0.000 [0.000, 0.000] & 0.798 [0.685, 0.894] \\ 
~ & Phi3 & 0.378 [0.285, 0.487] & 0.496 [0.387, 0.607] & 0.588 [0.479, 0.700] \\ 
\rowcolor{graybg} \multicolumn{5}{c}{Reasoning comprehensiveness} \\ 
MIMIC & Qwen-3B & 0.742 [0.635, 0.841] & 0.661 [0.537, 0.787] & 0.922 [0.868, 0.969] \\ 
~ & Llama & 0.421 [0.270, 0.571] & 0.000 [0.000, 0.000] & 0.743 [0.639, 0.839] \\ 
~ & Phi3 & 0.606 [0.508, 0.704] & 0.957 [0.918, 0.990] & 0.942 [0.892, 0.980] \\ 
\midrule
NIH-CXR & Qwen-3B & 0.724 [0.583, 0.862] & 0.790 [0.649, 0.912] & 0.916 [0.846, 0.972] \\ 
~ & Llama & 0.339 [0.186, 0.500] & 0.000 [0.000, 0.000] & 0.576 [0.425, 0.727] \\ 
~ & Phi3 & 0.768 [0.652, 0.867] & 0.933 [0.855, 0.988] & 0.976 [0.943, 1.000] \\
\midrule
MIDRC & Qwen-3B & 0.839 [0.721, 0.959] & 0.778 [0.659, 0.877] & 0.933 [0.888, 0.977] \\ 
~ & Llama & 0.403 [0.273, 0.532] & 0.000 [0.000, 0.000] & 0.805 [0.696, 0.903] \\ 
~ & Phi3 & 0.838 [0.750, 0.915] & 0.780 [0.678, 0.864] & 0.847 [0.747, 0.926] \\ 
\bottomrule
\end{tabular}
\end{table}

\newpage

\begin{table}[!phtb]
\caption{Reasoning performance comparison on baseline, SFT, SFT + GRPO using Gemini as evaluator.}
\label{stab:gemini}
\centering
\begin{tabular}{llccc}
\toprule
Dataset & Model & Baseline & SFT & SFT + GRPO \\
\midrule
\rowcolor{graybg} \multicolumn{5}{c}{Reasoning recall} \\ 
MIMIC & Qwen & 0.523 [0.446, 0.608] & 0.575 [0.481, 0.672] & \textbf{0.699 [0.630, 0.764]} \\ 
~ & Llama & 0.212 [0.134, 0.288] & 0.000 [0.000, 0.000] & \textbf{0.716 [0.621, 0.805]} \\ 
~ & Phi3 & 0.442 [0.365, 0.529] & \textbf{0.782 [0.719, 0.836]} & 0.778 [0.710, 0.840] \\ 
\midrule
NIH-CXR & Qwen & 0.681 [0.575, 0.768] & 0.767 [0.624, 0.895] & \textbf{0.892 [0.831, 0.943]} \\ 
~ & Llama & 0.250 [0.155, 0.345] & 0.000 [0.000, 0.000] & \textbf{0.575 [0.436, 0.718]} \\ 
~ & Phi3 & 0.625 [0.545, 0.702] & 0.750 [0.636, 0.846] & \textbf{0.842 [0.768, 0.907]} \\ 
\midrule
MIDRC & Qwen & 0.474 [0.413, 0.533] & 0.529 [0.417, 0.640] & \textbf{0.807 [0.746, 0.870]} \\ 
~ & Llama & 0.134 [0.078, 0.198] & 0.000 [0.000, 0.000] & \textbf{0.782 [0.670, 0.878]} \\ 
~ & Phi3 & 0.555 [0.469, 0.642] & \textbf{0.647 [0.529, 0.768]} & 0.630 [0.518, 0.729] \\ 
\rowcolor{graybg} \multicolumn{5}{c}{Reasoning comprehensiveness} \\ 
MIMIC & Qwen & 0.726 [0.585, 0.846] & 0.677 [0.548, 0.805] & \textbf{0.852 [0.784, 0.910]} \\ 
~ & Llama & 0.553 [0.385, 0.714] & 0.000 [0.000, 0.000] & \textbf{0.772 [0.674, 0.865]} \\ 
~ & Phi3 & 0.679 [0.576, 0.771] & 0.914 [0.862, 0.965] & \textbf{0.961 [0.922, 0.993]} \\ 
\midrule
NIH-CXR & Qwen & 0.810 [0.667, 0.930] & 0.820 [0.690, 0.937] & \textbf{0.981 [0.950, 1.000]} \\ 
~ & Llama & 0.373 [0.229, 0.533] & 0.000 [0.000, 0.000] & \textbf{0.568 [0.417, 0.731]} \\ 
~ & Phi3 & 0.859 [0.761, 0.948] & 0.933 [0.861, 0.987] & \textbf{0.981 [0.960, 1.000]} \\
\midrule
MIDRC & Qwen & 0.946 [0.886, 1.000] & 0.790 [0.654, 0.911] & \textbf{0.992 [0.985, 1.000]} \\ 
~ & Llama & 0.478 [0.328, 0.620] & 0.000 [0.000, 0.000] & \textbf{0.797 [0.687, 0.893]} \\ 
~ & Phi3 & 0.851 [0.771, 0.922] & 0.793 [0.681, 0.884] & \textbf{0.859 [0.773, 0.932]} \\ 
\bottomrule
\end{tabular}
\end{table}

\newpage

\begin{table}[!phtb]
\caption{Comparing GRPO and additional SFT on a small amount of reasoning data. SFT on reasoning only is fine-tuned on 200 samples.}
\label{stab:comparing}
\centering
\begin{tabular}{llcc}
\toprule
Dataset & Model & \makecell[c]{SFT disease\\ + GRPO} & \makecell[c]{SFT disease \\+ SFT reasoning} \\ 
\midrule
\rowcolor{graybg} \multicolumn{4}{c}{F1}\\ 
MIMIC & Qwen & 0.729 [0.665, 0.786] & 0.652 [0.564, 0.724] \\ 
~ & Llama & 0.826 [0.775, 0.871] & 0.805 [0.758, 0.848] \\ 
~ & Phi3 & 0.712 [0.651, 0.773] & 0.689 [0.626, 0.754] \\ 
\midrule
NIH-CXR & Qwen & 0.824 [0.752, 0.880] & 0.836 [0.780, 0.888] \\ 
~ & Llama & 0.862 [0.813, 0.903] & 0.895 [0.842, 0.908] \\ 
~ & Phi3 & 0.689 [0.617, 0.745] & 0.653 [0.592, 0.705] \\ 
\midrule
MIDRC & Qwen & 0.661 [0.562, 0.748] & 0.792 [0.738, 0.846] \\ 
~ & Llama & 0.967 [0.942, 0.987] & 0.873 [0.838, 0.903] \\ 
~ & Phi3 & 0.535 [0.396, 0.610] & 0.562 [0.415, 0.583] \\ 
\rowcolor{graybg} \multicolumn{4}{c}{Reasoning recall}\\ 
MIMIC & Qwen & 0.725 [0.660, 0.793] & 0.751 [0.676, 0.827] \\ 
~ & Llama & 0.647 [0.556, 0.739] & 0.763 [0.681, 0.836] \\ 
~ & Phi3 & 0.714 [0.644, 0.785] & 0.746 [0.679, 0.812] \\
\midrule
NIH-CXR & Qwen & 0.792 [0.713, 0.870] & 0.833 [0.725, 0.924] \\ 
~ & Llama & 0.567 [0.430, 0.702] & 0.650 [0.512, 0.779] \\ 
~ & Phi3 & 0.692 [0.606, 0.773] & 0.754 [0.672, 0.835] \\ 
\midrule
MIDRC & Qwen & 0.664 [0.559, 0.750] & 0.792 [0.738, 0.846] \\ 
~ & Llama & 0.798 [0.685, 0.894] & 0.871 [0.838, 0.903] \\ 
~ & Phi3 & 0.588 [0.479, 0.700] & 0.538 [0.415, 0.583] \\ 
\rowcolor{graybg} \multicolumn{4}{c}{Reasoning comprehensiveness}\\ 
MIMIC & Qwen & 0.922 [0.868, 0.969] & 0.825 [0.756, 0.887] \\ 
~ & Llama & 0.743 [0.639, 0.839] & 0.786 [0.697, 0.857] \\ 
~ & Phi3 & 0.942 [0.892, 0.980] & 0.946 [0.898, 0.984] \\ 
\midrule
NIH-CXR & Qwen & 0.916 [0.846, 0.972] & 0.934 [0.859, 0.983] \\ 
~ & Llama & 0.576 [0.425, 0.727] & 0.667 [0.508, 0.806] \\ 
~ & Phi3 & 0.976 [0.943, 1.000] & 0.976 [0.934, 1.000] \\ 
\midrule
MIDRC & Qwen & 0.933 [0.888, 0.977] & 0.948 [0.910, 0.983] \\ 
~ & Llama & 0.805 [0.696, 0.903] & 0.640 [0.493, 0.766] \\ 
~ & Phi3 & 0.847 [0.747, 0.926] & 0.636 [0.500, 0.753] \\ 
\bottomrule
\end{tabular}
\end{table}

\newpage
\begin{prompt}[Gemini evaluation reasoning script]
\label{prompt:gemini}
\begin{lstlisting}[language=prompt]
There is an AI assistant that tend to extract diseases from radiology report. You are given the report, the reasoning
given by the assistant and the result list given by the assistant. Your task is to evaluate whether the AI assistant is doing
a correct job.

********* Instructions:
1. The output is a list of formatted structures, each element in the list has the components of: 'phrase', 
'whether supported by report', 'target diseases'.
2. For the 'phrase' component, your task is to divide the reasoning into semantically independent part. Each part will lead to 
one structured element returned, and the part will be the 'phrase' component.
3. 'whether_supported_by_report' is a boolean to evaluate whether the 'phrase' is supported by the report.
4. 'target_diseases': which condition (including 'Support Devices')  this phrase is intended to target. 
    Target disease must be in one of the list: ['Atelectasis', 'Cardiomegaly', 'Consolidation', 'Edema', 
        'Enlarged Cardiomediastinum', 'Fracture', 'Lung Lesion', 'Lung Opacity', 'Pleural Effusion', 'Pleural Other', 
        'Pneumonia', 'Pneumothorax', 'Support Devices'], or the target disease is in the result list.
5. Each phrase can target to a several diseases or even no disease. So 'target diseases' are a list of candidate disease this 
    'phrase' is targeting for.
6. For target disease, especially pay attention to 'Support Devices' since that one has to be inferred from the text.
7. Target disease can also be those coming from the result list.
8. If there is no reasoning or the reasoning is clearly just repeating a list or not making sense, return an empty list.

********** Example
    ## Report
        FINDINGS: A cluster of heterogeneous opacities in the right lower lung has 
        has continued to grow since ___. 
        Otherwise, the lungs are clear. Moderate cardiomegaly, including severe left
        atrial enlargement is chronic; there is no pulmonary vascular congestion or
        edema. The thoracic aorta is heavily calcified.  There may be a new small,
        right pleural effusions or pneumothorax.
        IMPRESSION: Slowly progressive chronic right pneumonia, could be exogenous
        lipoid pneumonia, but tuberculosis is in the differential.  CT scanning
        recommended.  Nurse ___ and I discussed the findings and their
        clinical significance by telephone at the time of dictation.
    ## Reasoning
        According to the report, there is a cluster of heterogeneous opacities in the right lower lung.
        A pneumonia has developed.  There are bilateral pleural effusions, one on the right and one on the left.
        Additionally, there may be a small pneumothorax at the right lung base.  Supportive devices are mentioned, 
        like a chest tube, but not specified.
    ## Result list
        ['Pneumonia', 'Pleural Effusion', 'Pneumothorax', 'Support Devices']    
    
    ###### Expected Logical Output (The model will format this as JSON automatically):
    - Phrase: "According to the report, there is a cluster of heterogeneous opacities..." -> Supported: True, Target: ['Lung Opacity']
    - Phrase: "A pneumonia has developed" -> Supported: True, Target: ['Pneumonia']
    - Phrase: "There are bilateral pleural effusions..." -> Supported: False, Target: ['Pleural Effusion']

********** Here is your input
    ## Report
        {note}
    ## Reasoning
        {reasoning}
    ## Result
        {result}
********** Give your output
\end{lstlisting}

\end{prompt}

\end{document}